\ifdefined\pdfminorversion\pdfminorversion=7\fi
\ifdefined\pdfsuppressptexinfo\pdfsuppressptexinfo=-1\fi
\documentclass[10pt,twocolumn,letterpaper]{article}
\usepackage[T1]{fontenc}

\usepackage[pagenumbers]{cvpr}

\usepackage{booktabs}
\usepackage{multirow}
\usepackage{array}
\usepackage{tabularx}
\usepackage{float}
\usepackage{makecell}
\usepackage{amsmath,amssymb}
\usepackage{graphicx}
\usepackage{xcolor}
\usepackage{colortbl}                                %
\definecolor{ours}{HTML}{EEF3FA}                    %
\usepackage{tikz}
\usetikzlibrary{arrows.meta,positioning,calc,fit,backgrounds}
\DeclareRobustCommand{\keycap}[1]{\tikz[baseline=(k.base)]{\node[inner xsep=2.1pt,inner ysep=1.1pt,minimum width=2.2ex,minimum height=2.15ex,font=\fontfamily{FiraSans-TLF}\fontseries{sb}\fontsize{6.4}{7}\selectfont,text=black!72](k){#1};\begin{scope}[on background layer]\fill[black!38,rounded corners=1.8pt]([yshift=-0.7pt]k.south west)rectangle([yshift=-0.7pt]k.north east);\shade[top color=white,bottom color=black!10,draw=black!40,line width=0.3pt,rounded corners=1.8pt](k.south west)rectangle(k.north east);\end{scope}}}  
\newcommand{\prelim}[1]{#1}  
 
\newcommand{\na}{\textcolor{gray}{--}}              

\newcommand{\figorstub}[2]{\IfFileExists{#2}{\includegraphics[width=#1]{#2}}{\fbox{\parbox{#1}{\centering\small figure pending: \texttt{\detokenize{#2}}}}}}
\newcommand{\interval}[2]{{\scriptsize[#1, #2]}}

\newcommand{\Phat}{\ensuremath{\hat P}}
\DeclareMathOperator*{\argmax}{arg\,max}

\usepackage{microtype}
\renewcommand{\paragraph}[1]{\noindent\textbf{#1}\ }   
\newif\ifshowyxy\showyxyfalse
\newcommand{\yxy}[1]{\ifshowyxy\textcolor{orange}{[yxy] #1}\else #1\fi}
\newif\iftasktwomain
\tasktwomaintrue   %
\newlength\fullfigwidth
\AtBeginDocument{\setlength\fullfigwidth{\textwidth}}   %
\newcommand\tasktwofigure{%
\begin{figure*}[t]
\centering
\includegraphics[width=\fullfigwidth]{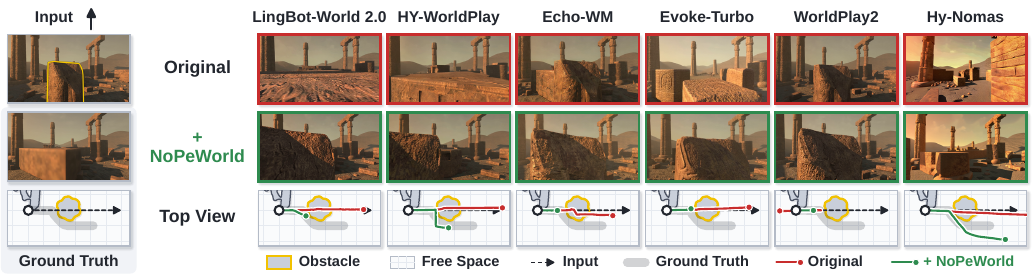}
\caption{\textbf{Task~2: the generated video.} One route, six generators, 4\,s after the hand-off; +~NoPeWorld in collision mode. Input: the hand-off frame (yellow: the obstacle) and the key held. Ground Truth: Unreal Engine. Top View: the reconstructed paths on the true map.}
\label{fig:task2}
\end{figure*}}

\usepackage{tcolorbox}
\usepackage{fontawesome5}
\usepackage{placeins}

\newtcbox{\tasktag}{
    on line,
    colback=black!5,
    colframe=black!25,
    boxrule=0.4pt,
    arc=2pt,
    boxsep=0pt,
    left=3pt,
    right=3pt,
    top=1pt,
    bottom=1pt,
 fontupper=\normalsize\bfseries
}

\definecolor{cvprblue}{rgb}{0.21,0.49,0.74}
\usepackage[pagebackref,breaklinks,colorlinks,allcolors=cvprblue]{hyperref}
\hypersetup{pdftitle={Non-penetrating World Models},pdfauthor={Yuhang Dai, Yinwei Wu, Ziyue Lin, Han Ji, Xingyi Yang}}

\newcommand{\projecturl}{https://nopeworld.github.io}
\newcommand{\codeurl}{https://github.com/yuhangdai427/NoPeWorld}
\newcommand{\weightsurl}{https://huggingface.co/PeterDAI/NoPeWorld}
\newcommand{\benchurl}{https://huggingface.co/datasets/PeterDAI/NoPeBench}
\newcommand{\linkicon}[1]{\raisebox{-0.15em}{\includegraphics[height=1em]{#1}}}
\newcommand{\linkline}{{\small
\linkicon{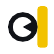}\,\href{\projecturl}{Project page}\enspace$\cdot$\enspace
{\faGithub}\,\href{\codeurl}{Code}\enspace$\cdot$\enspace
\linkicon{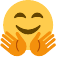}\,\href{\weightsurl}{Models}\enspace$\cdot$\enspace
\linkicon{figures/logo_hf.pdf}\,\href{\benchurl}{Data}}}

\title{\raisebox{-0.45em}{\includegraphics[height=1.6em]{figures/logo_nopeworld.pdf}}\hspace{0.35em}Non-penetrating World Models}

\usepackage{marvosym}
\newcommand{\labmark}{\raisebox{0.24em}{\includegraphics[height=1.05em]{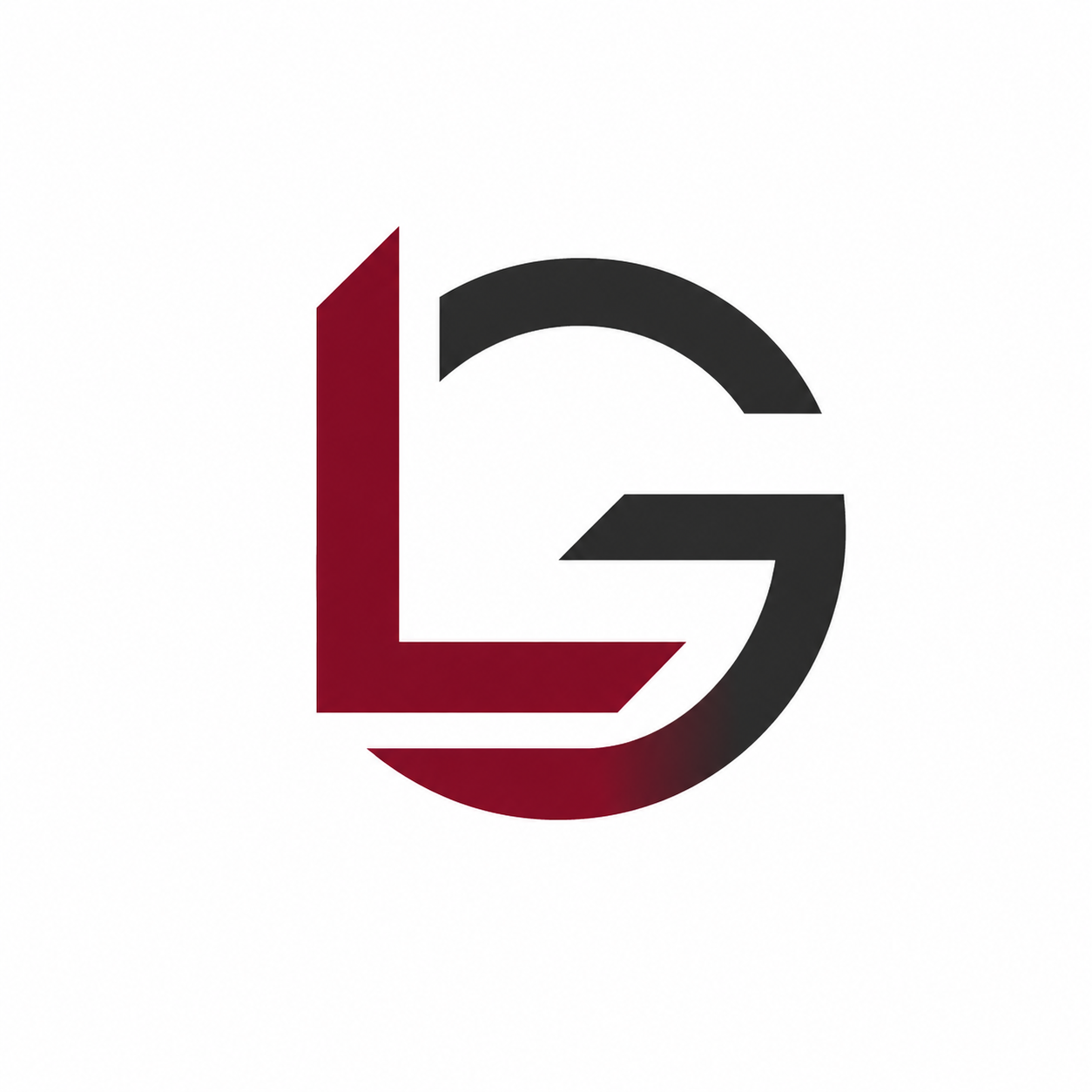}}}
\newcommand{\corrsymbol}[1]{\resizebox{!}{#1}{\Letter}}
\makeatletter
\newcommand{\corrauthor}{%
  \raisebox{0.48em}{\corrsymbol{0.52em}}%
  \g@addto@macro\@thanks{%
    \begingroup
    \renewcommand{\thefootnote}{\corrsymbol{3.8pt}}%
    \footnotetext[0]{Corresponding author.}%
    \endgroup}}
\makeatother
\newcommand{\polyuid}[1]{\href{mailto:#1@connect.polyu.hk}{#1}}
\newcommand{\authoremail}[1]{\href{mailto:#1}{\nolinkurl{#1}}}
\author{%
  Yuhang Dai\labmark\quad Yinwei Wu\labmark\quad
  Ziyue Lin\labmark\quad Han Ji\labmark\quad
  Xingyi Yang\labmark\corrauthor\\
  \labmark\,Learning \& Generation Lab, The Hong Kong Polytechnic University\\
  Hung Hom, Kowloon, Hong Kong\\
  {\small\ttfamily \{\polyuid{yu-hang.dai}, \polyuid{yinwei.wu}, \polyuid{robin.lin}, \polyuid{jihan.ji}\}@connect.polyu.hk}\\
  {\small\ttfamily \authoremail{xingyi.yang@polyu.edu.hk}}%
}

\begin{document}

\twocolumn[{%
\renewcommand\twocolumn[1][]{#1}%
\maketitle
\begin{center}
\vspace*{-20pt}\linkline\\[18pt]
\includegraphics[width=\fullfigwidth]{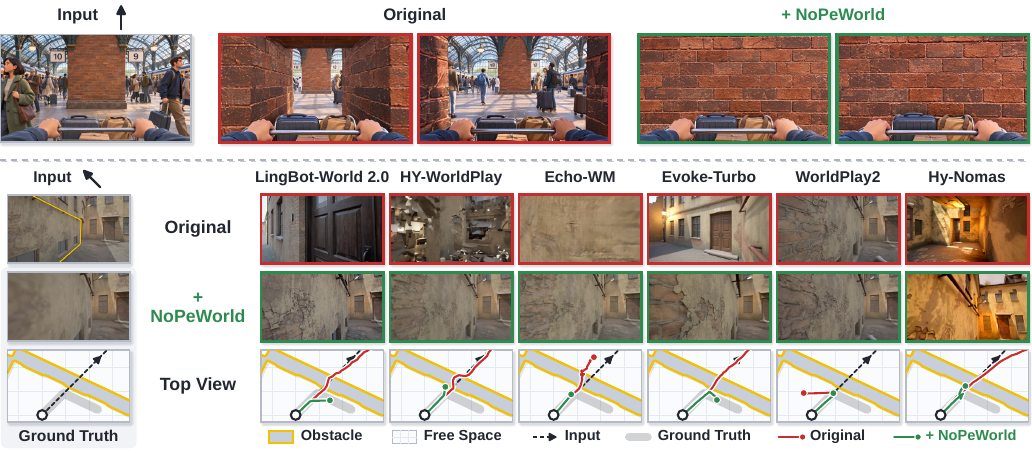}
\captionsetup{hypcap=false}%
\captionof{figure}{\textbf{Video world models walk through walls.} LingBot-World 2.0, \keycap{W} held at the pillar between platforms 9 and 10: Original walks through; +~NoPeWorld stops at the bricks. Six models on a NoPeBench route, \keycap{W}\,+\,\keycap{A} held (yellow: the wall; Top View: the true map): Original, five pass through, WorldPlay2 backs away; +~NoPeWorld, all stop at the wall, where the Ground Truth (Unreal Engine) slides.}
\label{fig:teaser}
\end{center}
}]
{\renewcommand{\thefootnote}{\corrsymbol{3.8pt}}\footnotetext[0]{Corresponding author.}}
\begin{abstract}
\yxy{We study a new problem of preventing object penetration in interactive video world models. To understand how existing models handle this, we introduce \textbf{NoPeBench}, a benchmark that evaluates how world models respond to physical obstacles. Our evaluation reveals that none of the open-source models can reliably prevent such failures.
Surprisingly, however, these models already encode the information needed to predict such violations. We find that the hidden states of diffusion transformers (DiTs) can predict whether a candidate action will penetrate an obstacle. Building on this, we introduce \textbf{NoPeWorld}, a lightweight framework that detects and prevents penetration. Specifically, we train a small prediction head on top of a frozen DiT's hidden states to read out if a candidate action is safe or not. Unsafe actions are then replaced with safe ones that best preserve the original intent.
Despite its simplicity, NoPeWorld substantially reduces the physical violation rate, averaged over six video world models, from 74\% to 16\%. Remarkably, our method shows zero-shot generalization to open-domain imagery, third-person viewpoints, and moving obstacles.
We publicly release our code, the NoPeBench benchmark, and model weights at the links above.}
\end{abstract}

\section{Introduction}
\label{sec:intro}

In the world of \emph{Harry Potter, Platform~9\textthreequarters{}} allows travelers to pass through a solid wall and enter another world~\cite{rowling1997}. In reality, however, a wall cannot be crossed because it's solid. Today's world models~\cite{genie,gamengen,matrixgame2,worldplay,lingbot,lingbotv2,worldplay2,echowm,evoke,hynomas} may still behave as if every wall were Platform~9\textthreequarters{}. 
As shown in~\cref{fig:teaser}, when commanded to move into a wall, five of the six generators we evaluate pass through it, while the remaining one moves against the user's command. These failures reveal that current world models struggle to enforce the non-penetration constraint. 

\yxy{The underlying principle, however, is not new. In computer graphics, \emph{non-penetration} requires that solid objects cannot occupy the same space~\cite{baraff1989}. Simulators and game engines have long enforced this constraint through \emph{collision detection} and \emph{collision response}~\cite{moore1988}. They detect collisions and adjust the motion to prevent penetration while preserving the intended movement as much as possible.}

\yxy{Unfortunately, this mechanism is missing from today's video world models. Basically, they are designed to follow user commands, whether camera poses~\cite{lingbotv2,worldplay,echowm,evoke,hynomas} or key inputs~\cite{worldplay2}, without checking whether the motion is physically possible. Since their training videos rarely, if ever, show a command blocked by a wall, such a command is likely out of distribution. As a result, cameras may pass through walls, or obstacles simply disappear. Worse still, existing benchmarks largely use collision-free trajectories, making it difficult to evaluate how well world models handle non-penetration~\cite{worldmark,wbench,worldroam,iworld}.}

\yxy{To prevent such failures, two straightforward solutions seem promising at first glance. The \emph{first} is to simply prompt the model to stop or slide along obstacles. However, we find that such prompts have little effect on penetration. The \emph{second} is to use an external depth estimator~\cite{da3,valerant} to detect obstacles and stop unsafe motions. While effective, this requires \emph{post-hoc processing} with an additional network for every generated frame. Moreover, stopping alone may sacrifice the user's intended motion. In that sense, both methods have their own limitations.}

\yxy{Surprisingly, we find that the generator itself already contains information needed to prevent penetration. Prior studies have shown that video diffusion models encode 3D structure and physical plausibility~\cite{invisible,beyondsurface,probe3d}. We find that this knowledge extends to predicting penetration. Specifically, the hidden states of a diffusion transformer (DiT) can predict whether a commanded action will cause penetration, even before the next video chunk is generated. In short, \emph{the generator already knows where its walls are, but its output fails to act on that knowledge.}}

\begin{figure*}[t]
\centering
\figorstub{\fullfigwidth}{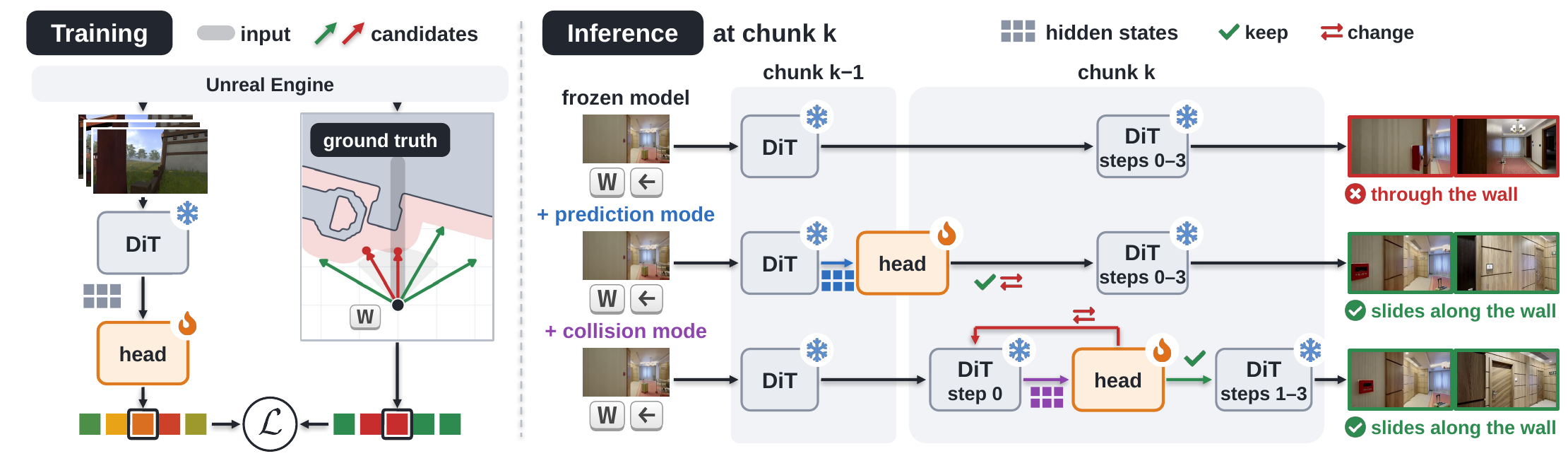}
\caption{\textbf{NoPeWorld.} Training: every candidate motion is swept through Unreal's true geometry (top view) and labelled; the head predicts the labels from the frozen DiT's states (outlined: keeping the input). Inference: raw, the model enters the wall; in prediction mode the head sets the motion between chunks $k{-}1$ and $k$; in collision mode it changes an unsafe motion inside chunk $k$ and reruns step~0.}
\label{fig:arch}
\end{figure*}

\yxy{Building on this observation, we introduce \textbf{NoPeWorld}, a lightweight framework that detects and prevents penetration in video world models. The rough idea is that rather than relying on external tools, we directly read the generator's hidden states to detect unsafe actions~(\cref{fig:arch}). Specifically, we train a small prediction head on a frozen DiT to predict collision risks and distances to obstacles. The head is trained using ground-truth distance and collision labels from Unreal Engine simulations. During inference, unsafe motions are replaced with safe alternatives that best preserve the user's intent, while safe motions remain unchanged.}

\yxy{We achieve this through two detection modes. In \emph{collision mode}, the head reads the initial denoising features of the current chunk to detect potential penetration. In \emph{prediction mode}, it uses features from the previous chunk to predict penetration in the upcoming chunk.}

A new task, of course, requires a new benchmark. To evaluate this, we introduce \textbf{NoPeBench}, a benchmark that deliberately challenges world models with obstacles~(\cref{sec:benchmark}). It consists of three levels of tasks. \tasktag{Task 1} evaluates, on ground-truth histories, whether a method keeps safe commands and corrects those that would penetrate.
\tasktag{Task 2} evaluates whether generated videos respect
non-penetration using the ground-truth geometry of scenes from UnrealZoo~\cite{unrealzoo}.
\tasktag{Task 3} tests generalization to open-domain scenes
using vision-language model (VLM) as judges.

Our experiments show that \textbf{NoPeWorld robustly detects and prevents penetration} across diverse video world models.
On LingBot-World~2.0~\cite{lingbotv2}, NoPeWorld intercepts nearly all unsafe commands (\cref{tab:decisions}), which cuts violations from \prelim{89\%} to \prelim{4\%} and lifts the Safe-Intent Score, zero for any video with a violation, from near zero to about half (\cref{sec:results,tab:main}). Moreover, the heads transfer without retraining: trained only on first-person walks through static scenes from UnrealZoo, they reduce Task~3 violations on all six generators, on open-domain imagery, from a third-person viewpoint, and around moving obstacles.

Our contributions are fourfold.
(i) \underline{\emph{Problem}}: To the best of our knowledge, this is the first work to study the problem of penetration prevention in video world models. (ii) \underline{\emph{Method}}: We introduce NoPeWorld, which detects collisions by probing the hidden states of a frozen DiT and replaces unsafe motions with the safe motion that best preserves the user's intent. (iii) \underline{\emph{Insights}}: We show that frozen world generators encode collision information even before future frames are generated (\cref{sec:knows}). Such collision awareness can be exploited across visual domains, camera viewpoints, and moving obstacles without retraining. (iv) \underline{\emph{Benchmark}}: We collect a new benchmark called NoPeBench. It evaluates whether video world models can prevent penetration when given commands that would otherwise lead to collisions.

\section{Related work}
\label{sec:related}

\paragraph{Interactive video world models.}
Interactive video generation has progressed from early action-conditioned predictors~\cite{genie,gamengen,diamond} to diffusion transformers~\cite{dit,wan} that denoise video chunk by chunk under camera-pose or keystroke control~\cite{cameractrl,film,matrixgame,matrixgame2,worldplay,lingbot,lingbotv2,echowm,evoke,forgewm,worldplay2,hynomas}. Because actions condition denoising in an open loop without scene geometry, these models ``do not explicitly enforce collision constraints''~\cite{valerant}. Prompting with obstacle descriptions fails to prevent clipping, while guarding decoded frames with an external depth estimator~\cite{da3,valerant} requires VAE decoding and auxiliary 3D inference at every chunk and only halts motion (\cref{sec:setup,sec:cost}). NoPeWorld instead keeps the generator frozen and enforces collision-aware control directly inside the DiT loop (\cref{sec:method}).

\paragraph{Collision handling in computer graphics.}
Computer graphics and mechanics have enforced solid non-penetration since the 1980s through collision impulses~\cite{moore1988,hahn1988}, unilateral contact constraints~\cite{lotstedt1982,lotstedt1984,moreau1988,baraff1989,baraff1994,stewart1996}, and penalty potentials~\cite{terzopoulos1987}, later extended to continuous swept collision detection against thin-wall tunnelling~\cite{redon2002} and intersection-free solvers~\cite{li2020ipc}. All of these methods require explicit 3D geometry and solve contact mechanics on known surfaces. Rather than simulating contact forces, we adapt two kinematic principles, continuous swept clearance and frictionless collide-and-slide projection (\cref{sec:problemdef}), to a video generator that stores no geometry, replacing unsafe commands with admissible stops or wall slides.

\paragraph{Internal representations of diffusion models.}
Intermediate features of image and video diffusion models capture semantic correspondence, depth, and 3D structure beyond surface appearance~\cite{baranchuk,dift,beyondsurface,probe3d}. Recent studies probe hidden states~\cite{alain} on intuitive-physics tasks~\cite{intphys} to show that video DiTs encode physical plausibility offline even when generated frames violate it~\cite{invisible}, or train a verifier on frozen DiT features to select among multiple sampled trajectories~\cite{seekingphysics}. Consistent with this view, we find that frozen DiT states already predict upcoming obstacle contact before the next chunk is synthesized, whereas the same readout on VAE latents or decoded pixels falls short (\cref{sec:knows}). Rather than offline probing or multi-sample rejection after generation, NoPeWorld queries frozen DiT tokens with a lightweight cross-attention head (\cref{sec:train}) to score and replace unsafe actions prior to chunk synthesis (\cref{sec:select}).

\paragraph{World model benchmarks.}
Physical video benchmarks~\cite{videophy,phygenbench,physicsiq} evaluate passive clips without user control, whereas interactive world-model benchmarks~\cite{iworld,worldinworld,acwmphys,worldmark,wbench,worldroam} focus on visual quality, memory, and action following along open paths: WorldMark~\cite{worldmark} removes wall-blocked trajectories, WBench~\cite{wbench} scores clipping as one VLM rubric item along mostly unobstructed routes (\cref{sec:benchmark}), and WorldRoamBench~\cite{worldroam} tests 58 clipping and 14 collision cases with VLMs. To evaluate non-penetration directly, NoPeBench (\cref{sec:benchmark}) commands every test route into an obstacle alongside open-path controls across teacher-forced (\emph{Task~1}), closed-loop (\emph{Task~2}), and open-domain (\emph{Task~3}) settings, pairing verified 3D reconstruction~\cite{vggtomega} and human-validated VLM judges with a hard-gated Safe-Intent Score inspired by NAVSIM~\cite{navsim} (\cref{sec:metrics}).

\section{NoPeWorld}
\label{sec:method}
\begin{figure*}[t]
\centering
\includegraphics[width=\fullfigwidth]{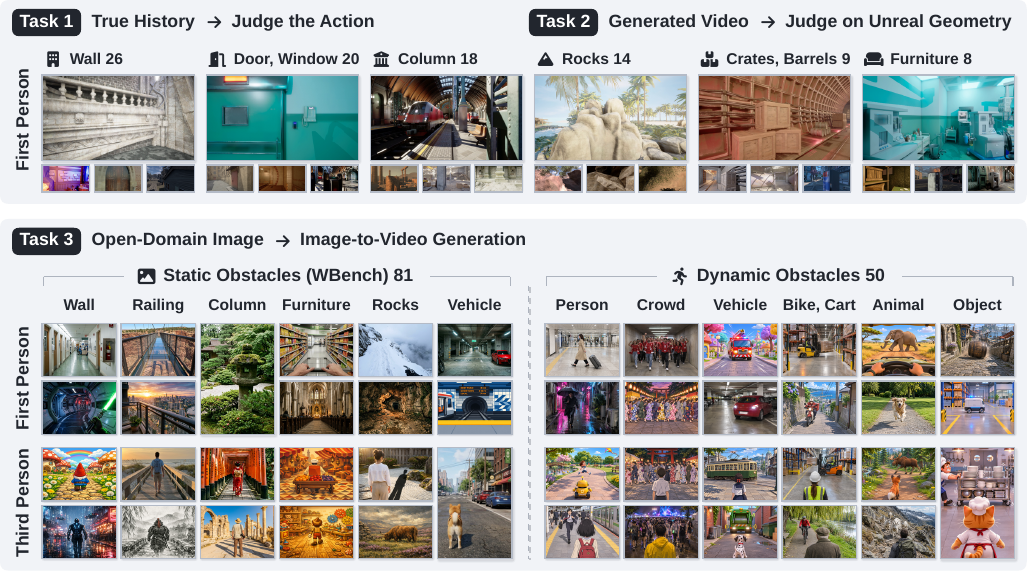}
\caption{\textbf{NoPeBench.} Tasks~1 and~2 use the same 100 routes in 32 held-out scenes from UnrealZoo~\cite{unrealzoo}, shown by what the camera runs into; Task~1 judges the action on the true history, Task~2 the generated video, on Unreal's geometry. Task~3 generates from open-domain images with static (WBench) and dynamic obstacles, in first and third person, judged by VLMs.}
\label{fig:benchmark}
\label{fig:coverage}
\end{figure*}

\subsection{Problem formulation}
\label{sec:problembench}
\label{sec:problemdef}
\label{sec:problem}

An interactive video world model synthesizes video chunk by chunk: given past frames $x_{<k}$ and a user command $u_k$ over the next $F$ latent frames, a diffusion transformer with parameters $\theta$ samples $x_k \sim p_\theta(x_k \mid x_{<k}, u_k)$. Because $u_k$ conditions generation in an open loop without geometric feedback, driving $u_k$ into a solid obstacle forces the model either to cross the obstacle or to hallucinate it away.

\noindent\textbf{Non-penetration in graphics}. Computer graphics enforces solid boundaries through two principles. First, ``A realistic simulation of rigid bodies demands that no two bodies inter-penetrate''~\cite{baraff1989}: a \emph{unilateral} constraint requires the signed obstacle gap $g\big(\gamma(\rho)\big) \ge 0$ along a swept path $\gamma(\rho)$ for $\rho \in [0,1]$~\cite{lotstedt1982,moreau1988,baraff1994}, tested over the full step so fast motion cannot \emph{tunnel} through thin walls~\cite{moore1988,redon2002}. Second, upon contact with surface normal $n = \nabla g$, a frictionless response that ``can `push' but not `pull'\,''~\cite{baraff1989,hahn1988} projects a commanded motion $u$ onto the closest admissible velocity, $v_S(u) = \operatorname*{arg\,min}_{v'\cdot n \ge 0}\|v'-u\|^2 = u - \min(0, u\cdot n)\,n$, halting head-on and sliding along oblique walls.

\noindent\textbf{Non-penetration in World Model}. Bringing both principles into a world model recasts chunk synthesis as constrained generation over chunk $x_k$ and executed action $a_k$:
\begin{align}
(x_k^\star,a_k^\star) = \argmax_{x_k,\,a_k}\;&\log p_\theta(x_k \mid x_{<k}, a_k) - \lambda\,\epsilon_S(a_k, u_k)^2 \nonumber\\
\text{s.t.}\;\;&\min_{\rho\in[0,1]} g_S\big(\gamma(\rho;\,a_k)\big) \ge 0 .
\label{eq:constrained}
\end{align}
Here $\min_{\rho\in[0,1]} g_S\big(\gamma(\rho;\,a_k)\big) \ge 0$ enforces continuous non-penetration against scene geometry $S$, $\lambda{>}0$ weighs intent against likelihood, and $\epsilon_S(a_k, u_k)$ is the distance of $a_k$'s displacement from the collide-and-slide response $v_S(u_k)$, relative to $|u_k|$, its cross-slide part weighted by $w{=}0.5$, allowing contact while penalizing stops or retreats. A generated video has a \emph{violation} if its trajectory enters solid geometry (\emph{penetration}), crosses an obstacle between frames (\emph{tunnelling}), or erases an obstacle on approach (\emph{vanishing}); obstacles may still move.

Freezing $\theta$ fixes the visual prior $p_\theta$ and leaves $a_k$ as the sole optimization variable; assuming that the generator renders the executed motion within the scene it has already shown, which Task~2 tests, the likelihood term of \cref{eq:constrained} no longer separates admissible actions, and a non-penetrating $a_k$ yields a non-penetrating video. Sweeping a collision disc ($r{=}0.25$\,m) over a discrete candidate set $\mathcal{C}$ of 34 motions (the raw input $c_0{=}u_k$, a stop, shortened steps, and turns up to $90^\circ$) with margin $m{=}0.05$\,m reduces \cref{eq:constrained} to
\begin{equation}
a_k^\star = \operatorname*{arg\,min}_{c_i\in\mathcal{C}}\ \epsilon_S(c_i,u_k)^2 \quad \text{s.t.}\quad g_S(c_i) \ge m,
\label{eq:discrete}
\end{equation}
with $g_S(c_i) \triangleq \min_{\rho\in[0,1]} g_S\big(\gamma(\rho;\,c_i)\big)$, the margin tightening ${\ge}\,0$ to ${\ge}\,m$ ($u_k$ is \emph{safe} if admissible and \emph{unsafe} otherwise) and $x_k^\star \sim p_\theta(\cdot \mid x_{<k}, a_k^\star)$ drawn by the generator's sampler. Since a video generator stores no explicit geometry $S$, neither $g_S(c_i)$ nor $\epsilon_S(c_i, u_k)$ can be evaluated at test time: NoPeWorld learns both terms of \cref{eq:discrete}, the constraint and the intent, from the DiT's hidden states, before any frame of the chunk is drawn.

\subsection{Constraint-aware training}
\label{sec:train}
\label{sec:head}

Prior to synthesizing chunk $k$, intermediate hidden states $H$ of a frozen video DiT $f_\theta$ already encode 3D scene structure and impending contact, whereas identical probes on VAE latents or decoded pixels fall well short (\cref{sec:knows,fig:headpos}). Building on this property, \textbf{NoPeWorld} (\textbf{No}n-\textbf{Pe}netrating \textbf{World}) trains a compact prediction head (weights $\phi$, under 2M parameters) on mid-to-deep DiT blocks while keeping every weight of $\theta$ frozen (\cref{tab:generators}; \cref{fig:arch}, Training, left). A token projector $\psi_H$ compresses all tapped tokens to width 128 without spatial pooling; candidate $c_i \in \mathcal{C}$ (and raw command $u_k$ when $H$ precedes $u_k$) is encoded by the generator's own frozen control encoder $E$ and projected onto the same token grid via $\psi_c$. A learned query $q$ then attends over the action-conditioned state tokens $\tilde H_i$ via two cross-attention blocks:
\begin{equation}
\begin{aligned}
(\ell_i, \hat g_i, \hat A_i) &= \mathrm{CrossAttn}_\phi\big(q,\ \tilde H_i\big),\\
\tilde H_i &= \psi_H(H) + \psi_c\big(E(c_i), E(u_k)\big),
\end{aligned}
\label{eq:head}
\end{equation}
predicting a safety logit $\ell_i$, swept clearance $\hat g_i$, and intent score $\hat A_i$ per candidate (details in \cref{s:head}).

To supervise the head on the unobserved terms of \cref{eq:discrete}, we replay 729 Unreal Engine~\cite{unrealengine} videos across 44 training scenes from UnrealZoo~\cite{unrealzoo} (disjoint from all benchmark scenes) through the frozen DiT via teacher forcing (\cref{fig:arch}, Training, left). On the simulator's true occupancy map (\cref{fig:arch}, Training, right), sweeping the collision disc gives the clipped clearance $g_i = \mathrm{clip}\big(g_S(c_i), -0.25, 1\big)$\,m, while comparing $c_i$ against the simulator's collide-and-slide displacement $v_S(u_k)$ gives the intent error $\epsilon_i = \epsilon_S(c_i, u_k)$ of \cref{eq:discrete}. Because the hard step $\mathbb{I}[g_S(c_i) \ge m]$ is discontinuous, we form smooth safety and intent targets ($s{=}0.5$):
\begin{equation}
P_i = \Phi\Big(\frac{g_i - m}{0.03\,\text{m}}\Big), \quad A_i = e^{-\epsilon_i^2/(2s^2)}, \quad R_i = P_i A_i,
\label{eq:labels}
\end{equation}
with standard normal CDF $\Phi$. Here $P_i$ softens the constraint of \cref{eq:discrete}, $g_S(c_i) \ge m$, $A_i$ maps its intent penalty $\epsilon_i^2$ monotonically into $(0,1]$, and $R_i$ ranks candidates that are both safe and intent-aligned. We train $\phi$ via multi-task supervision:
\begin{align}
\mathcal{L}(\phi) &= \sum\nolimits_i \omega_i\Big[\mathrm{BCE}\big(\sigma(\ell_i),P_i\big) + \frac{1}{2}\,\mathrm{Huber}(\hat g_i,g_i) \nonumber\\
&\qquad + \mathrm{BCE}(\hat A_i,A_i)\Big] + \mathrm{KL}\big(\pi(R)\,\|\,\pi(\hat R)\big),
\label{eq:loss}
\end{align}
so the head learns the constraint ($P_i$, with dense Huber regression on $g_i$ where $P_i$ saturates) and the objective ($A_i$) of \cref{eq:discrete}, plus a listwise ranking over $\mathcal{C}$ with logistic $\sigma$, $\hat R_i = \sigma(\ell_i)\hat A_i$ and $\pi(R)_i \propto R_i^{10}$. Weights $\omega_i$ sum to one, and chunk indices are reweighted to equalize unsafe rates so predictions rely on geometry, not time. A motion $c$ has \emph{regret} $\max_i R_i - R_c$.

\subsection{Two-mode closed-loop inference}
\label{sec:select}

At test time, plugging the head into the rollout solves \cref{eq:discrete} before any frame of chunk $k$ is decoded (\cref{fig:arch}, Inference). Given calibrated safety probabilities $\Phat_i = \sigma(\ell_i/T)$, candidates above threshold $\tau$ form the feasible set $\mathcal{S}_k = \{i : \Phat_i \ge \tau\}$, the learned constraint of \cref{eq:discrete}, and $u_k$ is projected onto the closest admissible motion:
\begin{equation}
a_k^\star = \begin{cases} u_k, & \text{if } 0\in \mathcal{S}_k,\ \max\limits_{i\in \mathcal{S}_k}\hat A_i \le \hat A_0 + \delta,\\[4pt] \argmax\limits_{i\in \mathcal{S}_k}\hat A_i, & \text{otherwise,}\end{cases}
\label{eq:select}
\end{equation}
defaulting to a stop when $\mathcal{S}_k = \emptyset$, with $(T, \tau, \delta)$ fixed on validation scenes (\cref{sec:selection}). In words, a safe input runs as is (\emph{kept}, $a_k^\star{=}u_k$), and an unsafe one is replaced by the safe motion closest to the intent (\emph{changed}, $a_k^\star{\neq}u_k$). NoPeWorld runs this loop in two modes that differ in where $H$ is tapped:
\begin{equation}
\begin{aligned}
H_{k-1}^{\text{com}} &= f_\theta(\hat z_{k-1},\, 0 \mid a_{k-1}^\star, M_{k-2}) &&\text{(prediction)},\\
H_k^{\text{step}} &= f_\theta(z_k^{(0)},\, t_0 \mid u_k, M_{k-1}) &&\text{(collision)}.
\end{aligned}
\label{eq:commit}
\end{equation}

\paragraph{Prediction mode (Pred).} After chunk $k{-}1$ is denoised into clean latent $\hat z_{k-1}$, the generator runs a noise-free \emph{commit pass} to write attention memory $M_{k-1}$. Pred reads this cached state $H_{k-1}^{\text{com}}$ with the incoming $u_k$, between chunks $k{-}1$ and $k$. Choosing $a_k^\star$ before chunk $k$ is denoised, Pred reruns no step and adds no DiT pass.

\paragraph{Collision mode (Coll).} Coll instead enters chunk $k$ and taps the first denoising pass $H_k^{\text{step}}$ (step~0, initial noise $z_k^{(0)}$ at level $t_0$), where $u_k$ already interacts with memory $M_{k-1}$, so the head needs no separate $E(u_k)$. If $u_k$ is kept, denoising proceeds directly from step~1 with no rerun. If $u_k$ is changed to $a_k^\star$, Coll reruns only step~0 once with $a_k^\star$ ($f_\theta(z_k^{(0)}, t_0 \mid a_k^\star, M_{k-1})$) and completes the remaining steps.

\begin{table}[t]
\centering\footnotesize
\caption{\textbf{Six generators, run frozen as released}, and the DiT blocks the head reads in each mode (--: none). $^\dagger$Concurrent work.}
\label{tab:generators}
\setlength{\tabcolsep}{1.9pt}
\begin{tabular}{@{}llll@{}}
\toprule
 & & DiT & Blocks read\\
Generator & Backbone & params & {\scriptsize Coll\,/\,Pred}\\
\midrule
LingBot-World 2.0~\cite{lingbotv2} & Wan2.2~\cite{wan} & 18.5B & 30,\,38,\,39\,/\,30\\
HY-WorldPlay~\cite{worldplay} & HunyuanVideo-1.5 & 8.6B & 27\,/\,33\\
Echo-WM~\cite{echowm} & LTX-2.3 & 19.8B & 42\,/\,30\\
Evoke-Turbo~\cite{evoke} & Helios & 14.3B & 20\,/\,\na\\
\addlinespace[2pt]
WorldPlay2$^\dagger$~\cite{worldplay2} & Wan2.2~\cite{wan} & $2{\times}17.5$B & 30\,/\,20\\
Hy-Nomas$^\dagger$~\cite{hynomas} & HunyuanVideo-1.5 & 8.6B & 33\,/\,33\\
\bottomrule
\end{tabular}
\end{table}

\begin{table}[t]
\centering\footnotesize
\caption{\textbf{The Task~3 judges against people} (269 majority-labeled videos, 44 violations). GPT-6 abstained on three videos (266 scored). $\kappa$: agreement with the Task~2 geometric evaluator. Shaded: the main table's two judges.}
\label{tab:evaluators}
\setlength{\tabcolsep}{4.5pt}
\begin{tabular}{@{}l l cccc@{}}
\toprule
Judge & & Recall$\uparrow$ & Precision$\uparrow$ & F1$\uparrow$ & $\kappa\uparrow$\\
\midrule
\rowcolor{ours}Claude Opus 5.5 & closed & \textbf{.91} & \underline{.78} & \underline{.84} & \textbf{.81}\\
\rowcolor{ours}GPT-6 Astra & closed & \underline{.81} & \textbf{.92} & \textbf{.86} & \textbf{.81}\\
DeepSeek V4.1 Flash & open & .55 & .71 & .62 & .67\\
Qwen3.8-27B & open & .59 & .70 & .64 & .68\\
\bottomrule
\end{tabular}
\end{table}

\begin{table}[t]
\centering\footnotesize
\caption{\textbf{Task~1: the decision on the true history} (100 routes; \% of decisions). The raw model misses 100 and changes 0. Bold: the better mode. --: no such mode. $^\dagger$Concurrent work.}
\label{tab:decisions}
\setlength{\tabcolsep}{3.2pt}
\begin{tabular*}{\columnwidth}{@{\extracolsep{\fill}}l cc cc ccc@{}}
\toprule
 & \multicolumn{2}{c}{Missed$\downarrow$} & \multicolumn{2}{c}{False$\downarrow$} & \multicolumn{3}{c}{Score$\uparrow$}\\
\cmidrule(lr){2-3}\cmidrule(lr){4-5}\cmidrule(l){6-8}
 & {\scriptsize Coll} & {\scriptsize Pred} & {\scriptsize Coll} & {\scriptsize Pred} & {\scriptsize Raw} & {\scriptsize Coll} & {\scriptsize Pred}\\
\midrule
LingBot-World 2.0 & 2.6 & \textbf{2.4} & \textbf{7.0} & 9.3 & 56.3 & \textbf{89.3} & 88.1 \\
HY-WorldPlay & \textbf{4.0} & 6.6 & \textbf{9.2} & 29.9 & 56.3 & \textbf{81.8} & 79.7 \\
Echo-WM & 3.3 & \textbf{2.5} & 19.2 & \textbf{12.7} & 51.2 & 83.7 & \textbf{84.6} \\
Evoke-Turbo & 11.5 & \na & 4.4 & \na & 47.2 & 79.7 & \na \\
WorldPlay2$^\dagger$ & \textbf{0.6} & 1.4 & \textbf{14.6} & 17.9 & 53.2 & \textbf{85.9} & 82.6 \\
Hy-Nomas$^\dagger$ & \textbf{1.5} & 5.7 & 49.4 & \textbf{31.1} & 55.7 & \textbf{76.7} & 72.7 \\
\bottomrule
\end{tabular*}
\end{table}

\begin{table}[!t]
\centering\footnotesize
\caption{\textbf{Main results} (\%). Task~2: SIS (\cref{eq:metrics}); Viol., routes whose video penetrates or erases the obstacle; Slide, share of the engine's slide; SIS 95\% CIs: \cref{s:tab-l2-full}. Task~3: violation rate. Overall: mean of SIS and the Task~3 share of safe videos with the input. \textbf{Best}, \underline{second} per generator. $^\dagger$Concurrent work.}
\label{tab:main}
\setlength{\tabcolsep}{3.2pt}
\renewcommand{\arraystretch}{1.065}
\begin{tabularx}{\columnwidth}{@{}X ccc cc c@{}}
\toprule
 & \multicolumn{3}{c}{Task 2: generated video} & \multicolumn{2}{c}{Task 3: violation} & \\
\cmidrule(lr){2-4}\cmidrule(lr){5-6}
Method & SIS$\uparrow$ & Viol.$\downarrow$ & Slide$\uparrow$ & Static$\downarrow$ & Dynamic$\downarrow$ & Overall$\uparrow$\\
\midrule
\multicolumn{7}{@{}l}{\textit{LingBot-World 2.0}~\cite{lingbotv2}}\\
\hspace{4pt}Raw model & 5 & 89 & 4 & 67 & 18 & 26.0 \\
\hspace{4pt}Text prompt & 4 & 94 & 3 & 70 & 22 & 25.1 \\
\hspace{4pt}External depth & 34 & \textbf{2} & 2 & \underline{30} & \underline{11} & 49.5 \\
\rowcolor{ours}\hspace{4pt}Ours (Coll) & \underline{50} & 7 & \underline{19} & \textbf{19} & \textbf{8} & \underline{61.6} \\
\rowcolor{ours}\hspace{4pt}Ours (Pred) & \textbf{56} & \underline{4} & \textbf{30} & \textbf{19} & \underline{11} & \textbf{66.7} \\
\specialrule{\lightrulewidth}{1pt}{1pt}
\multicolumn{7}{@{}l}{\textit{HY-WorldPlay}~\cite{worldplay}}\\
\hspace{4pt}Raw model & 5 & 86 & 2 & 22 & 11 & 37.9 \\
\hspace{4pt}Text prompt & 8 & 83 & 5 & 24 & 8 & 40.0 \\
\hspace{4pt}External depth & \underline{38} & \textbf{0} & 4 & \underline{15} & \underline{2} & 53.3 \\
\rowcolor{ours}\hspace{4pt}Ours (Coll) & \underline{38} & \underline{15} & \underline{14} & 16 & \textbf{0} & \underline{54.3} \\
\rowcolor{ours}\hspace{4pt}Ours (Pred) & \textbf{43} & 21 & \textbf{23} & \textbf{12} & \textbf{0} & \textbf{62.0} \\
\specialrule{\lightrulewidth}{1pt}{1pt}
\multicolumn{7}{@{}l}{\textit{Echo-WM}~\cite{echowm}}\\
\hspace{4pt}Raw model & 11 & 86 & 7 & 61 & 22 & 31.9 \\
\hspace{4pt}Text prompt & 12 & 86 & 10 & 60 & 27 & 31.8 \\
\hspace{4pt}External depth & 14 & \textbf{2} & 2 & 33 & 11 & 41.2 \\
\rowcolor{ours}\hspace{4pt}Ours (Coll) & \textbf{44} & \underline{3} & \textbf{18} & \underline{22} & \underline{8} & \textbf{62.7} \\
\rowcolor{ours}\hspace{4pt}Ours (Pred) & \underline{42} & 5 & \underline{12} & \textbf{15} & \textbf{7} & \underline{62.1} \\
\specialrule{\lightrulewidth}{1pt}{1pt}
\multicolumn{7}{@{}l}{\textit{Evoke-Turbo}~\cite{evoke}}\\
\hspace{4pt}Raw model & 8 & 77 & 5 & 48 & 22 & 34.3 \\
\hspace{4pt}Text prompt & 10 & 80 & \underline{7} & 50 & 23 & 34.3 \\
\hspace{4pt}External depth & \underline{11} & \textbf{4} & 1 & \underline{28} & \underline{7} & \underline{38.3} \\
\rowcolor{ours}\hspace{4pt}Ours (Coll) & \textbf{27} & \underline{30} & \textbf{13} & \textbf{17} & \textbf{1} & \textbf{55.0} \\
\specialrule{\lightrulewidth}{1pt}{1pt}
\multicolumn{7}{@{}l}{\textit{WorldPlay2$^\dagger$}~\cite{worldplay2}}\\
\hspace{4pt}Raw model & \underline{36} & 11 & \underline{31} & 16 & 14 & \underline{54.6} \\
\hspace{4pt}Text prompt & \textbf{43} & 9 & \textbf{36} & 12 & 8 & \textbf{60.5} \\
\hspace{4pt}External depth & 31 & \textbf{1} & 5 & 15 & \underline{7} & 51.7 \\
\rowcolor{ours}\hspace{4pt}Ours (Coll) & 25 & \textbf{1} & 6 & \textbf{7} & \textbf{4} & 52.1 \\
\rowcolor{ours}\hspace{4pt}Ours (Pred) & 29 & \underline{4} & 8 & \underline{10} & 8 & 50.7 \\
\specialrule{\lightrulewidth}{1pt}{1pt}
\multicolumn{7}{@{}l}{\textit{Hy-Nomas$^\dagger$}~\cite{hynomas}}\\
\hspace{4pt}Raw model & 2 & 95 & 0 & 52 & 47 & 17.8 \\
\hspace{4pt}Text prompt & 5 & 95 & 2 & 47 & 51 & 18.9 \\
\hspace{4pt}External depth & 20 & 50 & 11 & \textbf{29} & \underline{36} & 32.2 \\
\rowcolor{ours}\hspace{4pt}Ours (Coll) & \textbf{31} & \textbf{40} & \underline{17} & \underline{31} & \textbf{26} & \textbf{43.2} \\
\rowcolor{ours}\hspace{4pt}Ours (Pred) & \underline{29} & \underline{45} & \textbf{24} & 40 & 44 & \underline{36.8} \\
\bottomrule
\end{tabularx}
\end{table}

\iftasktwomain\tasktwofigure\fi
\section{Experiments}
\label{sec:experiments}

\subsection{NoPeBench}
\label{sec:benchmark}
\label{sec:construction}
\label{sec:tasks}

\textbf{NoPeBench} (\cref{fig:benchmark}) tests whether a world model enforces non-penetration without overriding safe user intent. Existing interactive benchmarks largely steer around obstacles: in WBench~\cite{wbench}, scripted trajectories drive into a visible obstacle in only 13 of 158 navigation cases by our count, recording clipping passively rather than testing contact. In NoPeBench every test command drives into a solid obstacle, and 50 unobstructed open-path routes check against unnecessary intervention, so both clipping and needless halting are penalized. All scenes, assets, and images are disjoint from training and validation (\cref{s:data}).

To separate perception from rollout drift and domain shift, NoPeBench has three progressive tasks. \tasktag{Task 1} isolates single-step decisions from generation drift: it replays ground-truth histories by teacher forcing on 100 collision routes in 32 held-out scenes from UnrealZoo~\cite{unrealzoo} and records the action chosen at each chunk. \tasktag{Task 2} runs 5\,s of obstacle-bound commands in closed loop on the same routes, from a pre-contact \emph{hand-off} on true frames. \tasktag{Task 3} starts from a single open-domain image without history (131 obstacle-bound images: 81 from WBench with static obstacles and 50 with moving ones, in first and third person).

\subsection{Evaluators and metrics}
\label{sec:metrics}
\label{sec:evaluator}
\begin{figure*}[t]
\centering
\includegraphics[width=\fullfigwidth]{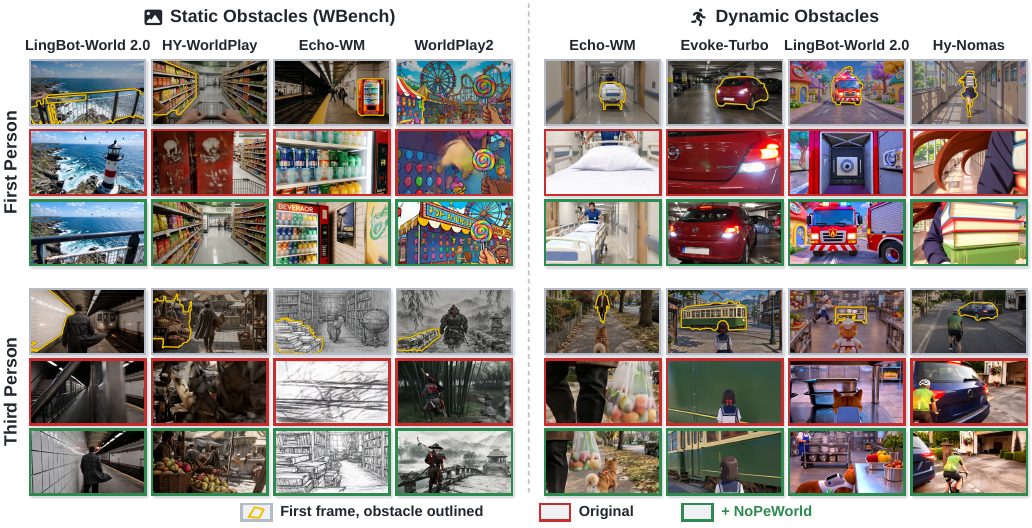}
\caption{\textbf{Task~3: open-domain images.} Static (WBench) and dynamic obstacles, first and third person, one generator per column. Rows: the first frame, obstacle outlined (SAM~2~\cite{sam2}); the original model; \mbox{+~NoPeWorld} (collision mode) at the same time. By VLM majority, every original shown violates and no \mbox{+~NoPeWorld} does.}
\label{fig:task3}
\end{figure*}

\paragraph{\tasktag{Task 1} Decisions on the true history.}
Task~1 asks whether a method changes unsafe inputs and keeps safe ones, on Unreal's occupancy map: \emph{Missed} (unsafe $u_k$ kept, $a_k^\star{=}u_k$), \emph{False} (safe $u_k$ changed), and the \emph{decision score} $100\,(1-\text{regret})$ (\cref{sec:train}), 100 at the reward-optimal motion.

\paragraph{\tasktag{Task 2} Geometric rollout evaluation.}
Task~2 scores non-penetration and intent in closed loop. VGGT-$\Omega$~\cite{vggtomega} reconstructs camera poses and depth, aligns the trajectory to Unreal's occupancy map (with SAM~2 tracking obstacle persistence), and flags a violation when penetration exceeds pose error (\cref{s:bench-eval}); it is right on 366 of 391 Unreal renders of known outcome and agrees with DA3~\cite{da3} on 3,350 generated videos ($\kappa{=}0.89$). Each rollout gets the \emph{Safe-Intent Score},
\begin{equation}
\mathrm{SIS}=\mathbf{1}[\text{safe}\wedge\text{with the input}]\cdot\exp\!\Big(-\frac{\mu^2+\nu^2}{2s^2}\Big),
\label{eq:metrics}
\end{equation}
alongside the \emph{violation rate} and \emph{Slide}, the share of the engine's wall slide a video reproduces. Mirroring $R{=}PA$ ($s{=}0.5$, \cref{eq:labels}), SIS is zero unless the rollout is \emph{safe} (\cref{sec:problemdef}) and \emph{with the input} (not stalling or opposing the command), while $\mu$ and $\nu$ penalize early stops and lost slide.

\paragraph{\tasktag{Task 3} VLM-judged open-domain evaluation.}
Without 3D ground truth, Task~3 relies on VLM judges checked against people and geometry. Of four VLMs judging each video under one fixed prompt, Claude Opus~5.5 and GPT-6 Astra agree most closely with the human reference labels and the Task~2 evaluator (\cref{tab:evaluators}); lacking a reference slide, Task~3 reports their mean \emph{violation rate}.

\subsection{Experimental setup}
\label{sec:setup}
\label{sec:five}
\label{sec:selection}

\paragraph{Models and configuration.}
One head architecture serves six frozen open-source generators with their released weights and samplers (\cref{tab:generators}), spanning four backbones, five driven by camera poses and one by keystrokes (WorldPlay2), by reusing each model's native control encoder; memoryless Evoke-Turbo supports Coll only. Readout blocks and $(T, \tau, \delta)$ are fixed on validation scenes (\cref{s:select}).

\paragraph{Compared methods.}
We compare four methods under identical histories, captions, and seeds: open-loop generation (\emph{Raw model}); a privileged \emph{Text prompt}, given the ground-truth obstacle name, side, and metric distance with instructions to stop or slide (\cref{s:baselines}); a halt-only \emph{External depth} guard, which decodes each chunk and halts via DA3~\cite{da3} whenever $u_k$ is swept unsafe; and \emph{NoPeWorld} (Coll and Pred inside the frozen DiT, \cref{sec:select}).

\subsection{Main results}
\label{sec:results}

\paragraph{\tasktag{Task 1} Single-step detection and action selection.}
On drift-free ground-truth histories, NoPeWorld intercepts unsafe commands and keeps most safe ones on all six generators (\cref{tab:decisions}). Both modes cut missed collisions from 100\% to 0.6--11.5\% and raise the decision score by 17--33 points; Pred nearly matches Coll, though it reads the DiT before $u_k$ enters. The better mode alters 4--15\% of safe inputs, 31.1\% on Hy-Nomas. Task~2 tests whether these decisions hold in closed loop.

\paragraph{\tasktag{Task 2} Closed-loop video generation on 3D geometry.}
In autoregressive rollouts, these decisions give safer videos that keep the intent: NoPeWorld has the best SIS and Overall on all five pose-driven generators (on HY-WorldPlay in prediction mode; collision mode ties \emph{External depth} on SIS), with up to 3.1$\times$ the SIS of \emph{External depth} (\cref{tab:main}; \iftasktwomain\cref{fig:teaser,fig:task2}\else\cref{fig:teaser}\fi). Their raw rollouts violate on 77--95\% of routes with near-zero SIS, and \emph{Text prompt} brings almost no gain. \emph{External depth} has the fewest violations on four of them, but only by halting: stuck at a wall it cannot slide along, it collapses \emph{Slide} and restricts the user's motion. Projecting unsafe $u_k$ onto admissible slides and stops (\cref{eq:select}), NoPeWorld cuts violations (from 74\% to 16\% averaged over all six in collision mode) while recovering part of the wall slide. WorldPlay2, the only keystroke-driven model, moves on its own: it natively retreats rather than penetrates, and a text prompt can steer it, unlike pose-driven models. NoPeWorld removes nearly all its remaining violations (11\% to 1\% in collision mode) and reduces retreats, but its stops take part of the native slide, leaving its SIS below the raw model's, the prompt's and \emph{External depth}'s. On the 50 open-path control routes, both modes neither stop early nor retreat in 92--100\% of the videos (\cref{s:l2-ctl}), so NoPeWorld rarely halts where nothing blocks the way.

\paragraph{\tasktag{Task 3} Zero-shot transfer to open-domain scenes.}
Is this ability tied to the training geometry? Though trained solely on teacher-forced first-person walks in static UnrealZoo scenes, NoPeWorld transfers to open-domain images, single-frame starts, third-person views and moving obstacles (\cref{tab:main,fig:task3}): both modes cut static and dynamic violations below the raw model, and to or below the text prompt, on every generator (static, LingBot-World~2.0: 67\% to 19\%). Coll matches or outperforms \emph{External depth} in 10 of the 12 generator-setting pairs, including all six with moving obstacles (Echo-WM: 8\% vs.\ 11\%); its two static losses are by 1--2 points. This suggests frozen DiTs encode transferable contact cues.

\section{Analysis}
\label{sec:analysis}

To see why reading the DiT works, we ask whether its states hold the coming contact before pixels exist (\cref{sec:knows}), which blocks carry it (\cref{sec:position}), and at what cost (\cref{sec:cost}).

\subsection{The DiT knows before it draws}
\label{sec:knows}

\paragraph{The contact is predictable.}
A frozen DiT holds the coming contact before its frames exist. We replay teacher-forced Unreal videos of 8 held-out scenes on LingBot-World~2.0, with labels from the true geometry, and sort decision states by chunks until the input becomes unsafe. A head separates them from safe states through $\Phat_i$ strongly for the chunk about to be generated and, one chunk earlier, before those frames are denoised, still far above a lookup that sees only the time; two chunks ahead much of the margin is lost (\cref{s:probe-results}). Across all six generators, the head first intervenes later on routes with a later time to contact (\cref{s:leadtime}).

\paragraph{DiT states beat VAE latents and pixels.} Only DiT states support low-regret choices (bottom of \cref{fig:headpos}): on them the head roughly halves the regret of always keeping the input, whereas on VAE latents or pixels it is no better, or hardly better, and on pixels it tells unsafe inputs apart no better than the chunk index does (AUC, \cref{s:tab-ablation}). The DiT has already done the work, and a head of under two million parameters can pick it up. Yet the same model, raw, walks through the obstacle on most routes (\cref{tab:main}): it knows, and does not act on what it knows, likely because a blocked command is out of distribution (\cref{sec:intro}); NoPeWorld makes it act by turning this knowledge into the choice of $a_k^\star$ (\cref{eq:select}).

\subsection{Where to read}
\label{sec:position}

\begin{figure}[t]
\centering

\includegraphics[width=\columnwidth]{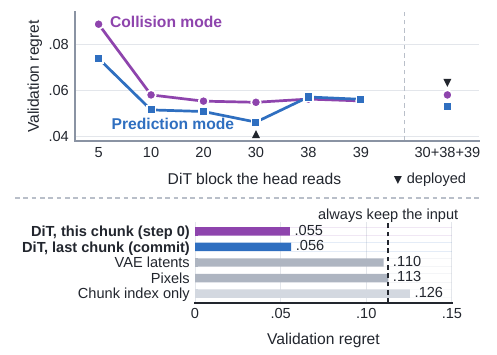}
\caption{\textbf{Where and what to read} (LingBot-World 2.0, validation regret, lower is better). Top: one DiT block, or blocks 30, 38 and 39, at step~0 (Coll) and at the previous commit pass (Pred); triangles: deployed heads. Bottom: block-39 DiT states, VAE latents, pixels and the chunk index alone; dashed: keeping the input.}
\label{fig:headpos}
\end{figure}

Where the head reads, our main design choice, turns out to be forgiving: on LingBot-World~2.0 the signal is spread over the middle and deep blocks rather than held by one layer, which likely helps one head design serve all six generators. Trained with $H$ (\cref{eq:head}) tapped at different blocks and scored on the 8 validation scenes, disjoint from training and benchmark scenes (top of \cref{fig:headpos}), the head serves the current chunk equally well from the tenth block on, within the spread between seeds; on the commit pass block 30 is lowest and the last two blocks somewhat higher. Only the shallowest is clearly worse on both, most of all at step~0, where its tokens are still near noise. At every block the commit pass matches step~0, so Pred loses nothing by reading a chunk early.

\subsection{Inference cost}
\label{sec:cost}

\begin{table}[t]
\centering\footnotesize
\caption{\textbf{Cost on LingBot-World~2.0} (s; 49 routes, 7 decisions; raw model 53.7\,s per route). Per decision: median, step~0 included for collision mode. Added: mean over the raw model on the same route; Pred: the head's own time.}
\label{tab:cost}
\setlength{\tabcolsep}{2.5pt}
\begin{tabular}{@{}lll@{}}
\toprule
Method & Per decision & Added per route\\
\midrule
\rowcolor{ours}Ours (Coll) & 2.11 (1.04 if kept) & $+$9.9\\
\rowcolor{ours}Ours (Pred) & 0.09 & $+$0.7\\
External depth & 2.21 (1.01 decode + 1.20 DA3) & $+$14.7\\
\bottomrule
\end{tabular}
\end{table}

Reading DiT states avoids pixel-space decoding and depth estimation (\cref{tab:cost}): per route, \emph{External depth} adds 14.7\,s by decoding and running DA3 per chunk, Coll 9.9\,s by rerunning step~0 only when $a_k^\star{\neq}u_k$, and Pred 0.7\,s, scoring all 34 candidates from the cached commit pass. Relative to the 53.7\,s raw route, Pred adds about 1\%, Coll 18\% and \emph{External depth} 27\%.

% Keep the accepted conclusion's paragraph and heading spacing as the references reflow.
\begingroup
\setlength{\parskip}{0pt}
\let\normalcvprsection\cvprsection
\patchcmd{\cvprsection}{-10pt plus -2pt minus -2pt}{-8.06604pt}{}{}
\section{Conclusion}
\label{sec:discussion}
\label{sec:conclusion}
Video world models inhabit a world where walls are optional. We formulated non-penetration for generators without geometry, requiring that the depicted trajectory never penetrate an obstacle and that no obstacle be erased, and introduced NoPeBench to evaluate it on ground-truth 3D geometry and on open-domain scenes. Rather than rely on external perception, NoPeWorld resolves collisions from within: because a frozen DiT already encodes impending contact before it synthesizes pixels, a small head suffices for collision detection and response, and transfers without retraining to new domains, viewpoints, and moving obstacles. Across six frozen generators, it cuts violations from 74\% to 16\%. Solid boundaries are only the first step; the frontier ahead is contact that pushes, carries, and shatters objects.

{
    \small
    \bibliographystyle{ieeenat_fullname}
    \let\originalbibsection\bibsection
    \renewcommand{\bibsection}{%
      \par\let\cvprsection\normalcvprsection
      {\normalsize
\section*{LLM Usage Statement}
The authors take full responsibility for all content, and all AI-assisted outputs were reviewed, edited, and approved by the authors. Claude Fable~5.2, Claude Opus~5.5, GPT-6 Astra, Gemini Carbon~4, and Moonshot models assisted in polishing author-written text, related work, and manuscript feedback. Claude Code assisted in implementing experiment scripts, the dataset pipeline, and diagrams and plots. OpenAI's GPT image generation produced synthetic starting images (\eg, in \cref{fig:teaser} and dynamic Task~3), whereas all video frames in \cref{fig:teaser,fig:task2,fig:task3} are outputs of the six evaluated world models (\cref{tab:generators}). Finally, Task~3 uses four VLMs as automated evaluators (\cref{sec:metrics}), validated against three human annotators (\cref{tab:evaluators}).
\par}%
      \originalbibsection}
    \bibliography{main}
}

\endgroup
\clearpage
\appendix
\renewcommand{\thesection}{S\arabic{section}}
\renewcommand{\theHsection}{S\arabic{section}}
\setcounter{table}{0}
\setcounter{figure}{0}
\renewcommand{\thetable}{S\arabic{table}}
\renewcommand{\theHtable}{S\arabic{table}}
\renewcommand{\thefigure}{S\arabic{figure}}
\renewcommand{\theHfigure}{S\arabic{figure}}
\makeatletter
\renewcommand\table[1][tbp]{\@float{table}[#1]}
\renewcommand\figure[1][tbp]{\@float{figure}[#1]}
\setcounter{dbltopnumber}{4}
\setcounter{topnumber}{3}
\setcounter{bottomnumber}{2}
\setcounter{totalnumber}{5}
\renewcommand{\topfraction}{0.9}
\renewcommand{\bottomfraction}{0.85}
\renewcommand{\textfraction}{0.07}
\renewcommand{\floatpagefraction}{0.8}
\renewcommand{\dblfloatpagefraction}{0.8}
\makeatother
\renewcommand{\dbltopfraction}{.94}
\twocolumn[{%
\centering{\Large\bfseries\thetitle\par}
\vspace{.5em}{\Large Supplementary Material\par}
\vspace{.9em}
\begin{minipage}{\textwidth}\captionsetup{type=figure}

\centering
\includegraphics[width=\textwidth]{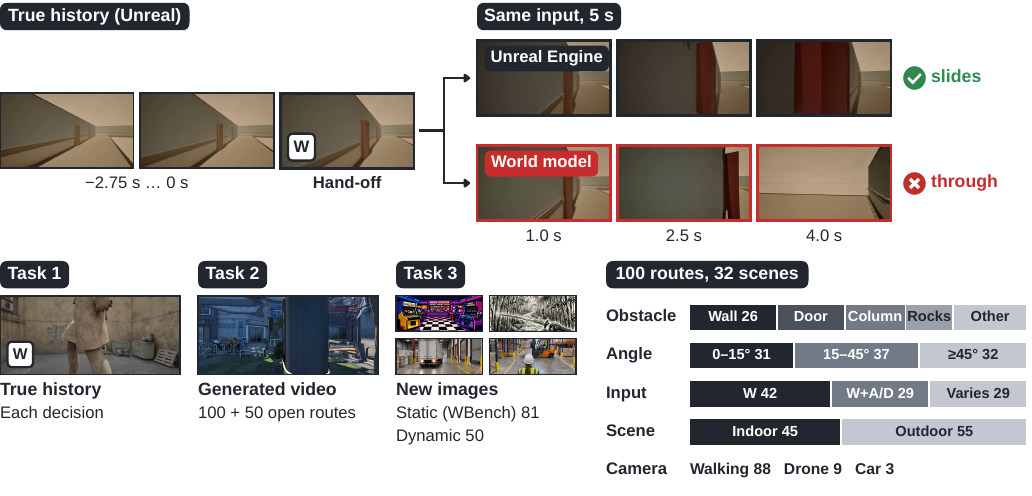}
\caption{\textbf{How a case of Tasks~1 and~2 is built} (a hospital corridor wall at 61\textdegree, input W). The true history in Unreal Engine runs up to the hand-off, and the same input then continues for 5\,s: Unreal Engine executes it with collision on and slides along the wall, while raw LingBot-World~2.0 passes through it. Below: what each task judges, and the distribution of obstacles, approach angles, inputs, scenes and cameras across the 100 routes.}
\label{s:fig-protocol}
\end{minipage}\par\vspace{6pt}}]
\raggedbottom
\begin{table*}[t]
\centering\footnotesize
\caption{\textbf{The evaluator's rule}, checked from top to bottom; the first row that applies decides. Thresholds at walking speed; $V$ is the commanded speed. In italics, the outcome category used in the result tables: \emph{through} is penetration or tunnelling, \emph{vanished} is vanishing, \emph{past} is walking on. At an oblique obstacle with room to slide, a camera that reaches the obstacle but falls short of the required slide and does not stop is labelled \emph{stopped, not slid}.}
\label{s:tab-verdicts}
\scriptsize
\setlength{\tabcolsep}{3pt}
\begin{tabularx}{\textwidth}{@{}>{\raggedright\arraybackslash}p{0.12\textwidth}>{\raggedright\arraybackslash}X>{\raggedright\arraybackslash}p{0.33\textwidth}@{}}
\toprule
Verdict & Condition & Origin of the threshold\\
\midrule
\multicolumn{3}{@{}l}{\emph{Violation} (any of the four)}\\
Penetration (\emph{through}) & camera centre deeper inside geometry than the video's noise floor, or in free space at least 0.10\,m from geometry where the free path from the hand-off camera exceeds 1.1$\times$ its own path $+$ 0.3\,m (through a thin wall), for $\ge2$ consecutive samples at 8\,fps; the noise floor is the video's own history error, clipped to [0.10, 0.30]\,m & 0.10\,m: position error near walls on renders with true poses, 95th percentile 0.061\,m within 0.6\,m, 0.117\,m within 0.3\,m with DA3, 0.063\,m and 0.096\,m with VGGT-Omega. 0.30\,m: the evaluator's position error on 214 Unreal reference executions (90th percentile of the per-video mean, 0.28\,m with DA3, 0.20\,m with VGGT-Omega)\\
Tunnelling (\emph{through}) & $D_t$ jumps within 0.125\,s to $\ge\max(2\,\text{m}, 3D_{t-1})$ from $D_{t-1}\le0.8$\,m, and at the pose extrapolated from two samples geometry fills the window centre or the camera would be inside or behind it & a depth ratio holds under a wrong scale; extrapolation tells a jump through geometry from the view opening past a pole or a wall's end\\
Vanishing (\emph{vanished}) & the same jump from $D_{t-1}>0.8$\,m with the obstacle still filling the window centre; or SAM~2 loses the obstacle while it is ahead, $>0.6$\,m away, off the image border, covering 1--50\% of the frame; either only if, at the extrapolated pose, the occupancy map still has eye-level geometry in at least half of the window centre within the same reach & a wall filling the frame gives the tracker no information, hence 50\%; a camera passing the end of an obstacle sees the background open up, and the map is empty there\\
Walking on (\emph{past}) & the centre has touched geometry and gets $\ge\max(1.5\,\text{m}, 0.375\,\text{s}\cdot V)$ further along the wall's inward normal than the Unreal reference execution ever does & going around with clearance never counts\\
\midrule
\multicolumn{3}{@{}l}{\emph{Motion falls short of the reference response} (progress 0)}\\
No response (\emph{no response}) & never more than 0.15\,m from the hand-off point & \\
Going back (\emph{reversed}, \emph{retreated}) & \emph{reversed}: $\ge0.3$\,m behind the hand-off point and ending behind it; \emph{retreated}: peak progress along the input minus end progress $\ge\max(0.3\,\text{m}, 0.1\,\text{s}\cdot V)$ & correctly judged reference renders: estimated retreat 99th percentile 0.17\,m with DA3, 0.09\,m with VGGT-Omega\\
Slow start (\emph{no response}, \emph{wrong direction}) & $<0.2V$ along the input in the first second and never at the obstacle; wrong direction if the end point is $>45$\textdegree{} off the input & \\
Early stop (\emph{early stop}, \emph{still moving}) & never at the obstacle (\emph{still moving} if it moved $>0.3$\,m in the last 2\,s) & a stop counts as at the obstacle within 0.6\,m\\
Turned away (\emph{turned away}) & at the obstacle, final heading $>45$\textdegree{} off the input's & correctly judged reference renders: 99th percentile 9.1\textdegree{} with DA3, 5.3\textdegree{} with VGGT-Omega\\
\midrule
\multicolumn{3}{@{}l}{\emph{Motion matches the reference response}}\\
Head-on (\emph{stopped}) & approach angle $<15$\textdegree: the camera reached the obstacle & the head-on bin of the routes\\
Oblique, room (\emph{slid}) & reference slides $>0.05$\,m along the wall: the camera slides along the tangent $\ge50$\% as far as the reference & \\
Oblique, no room (\emph{stopped}) & reference slide $\le0.05$\,m: the camera stops (last-second speed $<0.1V$) & where nothing can slide, a stop is the correct response\\
\bottomrule
\end{tabularx}
\end{table*}
\section{Benchmark (NoPeBench)}
\label{s:bench}

\subsection{Data sources and splits}
\label{s:data}
Every scene of Tasks~1 and~2 and of the heads' training comes from UnrealZoo~\cite{unrealzoo}, split by scene: the heads train on 729 videos of 44 scenes and are selected on held-out validation scenes (\cref{s:select}), while the benchmark routes lie in 32 separate test scenes (\cref{s:bench-cases}). Task~3 uses no engine: its static images are WBench~\cite{wbench} cases in first and third person, and its dynamic images are first frames made in WBench's manner with OpenAI's GPT image generation from scene descriptions we wrote (\cref{s:bench-l3}). The 32 test scenes share no asset pack with the training scenes. The 100 collision routes and 50 unobstructed control routes were fixed purely from scene geometry prior to model evaluation, and training and tuning used no Task~3 images.

\subsection{How cases are built}
\label{s:bench-cases}

\paragraph{Scenes and routes.}
\Cref{s:fig-protocol} shows one case from history to verdict. Cases are rendered in Unreal Engine~\cite{unrealengine}, in photo-realistic scenes of UnrealZoo~\cite{unrealzoo} driven through UnrealCV~\cite{unrealcv}, strictly disjoint from all training and validation scenes of the heads. A route is built backwards from a chosen spot on a wall or an object and an incidence-angle bin, so that the obstacle, the approach angle and the time to contact are controlled.

\paragraph{Hand-off and command.}
Each route starts with a ground-truth history clip up to a pre-contact \emph{hand-off} (about 1\,s before obstacle contact on 49 routes and 0.5--3\,s on the other 51), after which the model receives 5\,s of a persistent command: W, W+A or W+D, at 1.2\,m/s for walking, 4\,m/s for drones and 5\,m/s for cars. Every model starts from true frames up to the hand-off, in whole chunks of its own grid.

\paragraph{Routes.}
Tasks~1 and~2 share 100 routes in 32 held-out scenes (66 walls, 34 objects), chosen so that no two routes repeat the same scene, obstacle, input type and contact time; they run into 84 different obstacles, on foot and at drone and car speed. Glass, thin poles and moving bodies are excluded.

\subsection{The three tasks}
\label{s:tasks-full}

\paragraph{\tasktag{Task 1} The action, on the true history.}
\label{s:bench-l1}
Task~1 isolates the decision from the generator: the generator replays each route's real continuation by teacher forcing on its own chunk grid, and at every chunk we record the motion $a_k^\star$ the method would send, with every candidate labelled on the occupancy map as in training (\cref{eq:labels}). Task~1 has the two error rates of any detector: \emph{Missed}, the share of unsafe inputs ($P_0<0.5$) that are kept, and \emph{False}, the share of safe inputs that are changed. Candidate $c_i$ has reward $R_i=P_iA_i$ (the training target), and a motion $c$ has regret $\max_iR_i-R_c$. A label looks one chunk ahead, so it depends on the chunk length; we therefore compare Task~1 scores within each model.

\paragraph{\tasktag{Task 2} The generated video.}
\label{s:score-full}
On the same routes the model generates with the method in the loop, and a geometric evaluator places the camera path the video depicts on Unreal's occupancy map and gives it one verdict of \cref{sec:problemdef} (\cref{s:rules}); Unreal's own execution of the input, a stop or a slide, is the reference. The main paper's Safe-Intent Score (its \cref{eq:metrics}) is averaged over the 100 routes. A video is \emph{safe} if it has no violation, and \emph{with the input} if the camera does not go back, go the wrong way, turn away or stand still. $\mu$ is how many chunks of travel the camera stops short of the engine's contact, at its closest approach to the obstacle, with 0.3\,m allowed for position error; a chunk of travel is taken at the generator's own speed and chunk length. $\nu$ is the share of the engine's slide along the wall that the video misses, on the 65 routes where the engine slides at least 0.5\,m, and zero elsewhere. With $s{=}0.5$, the width of the intent label (\cref{eq:labels}), each unit of shortfall, one chunk short or no slide where the engine slides, multiplies the score by $e^{-2}$.

\paragraph{\tasktag{Task 3} Image-to-video generation from open-domain images.}
\label{s:bench-l3}
Task~3 leaves the training domain: the model generates video from a single open-domain image, not rendered in Unreal, in first or third person, with an input of ours that drives into a visible obstacle. The \emph{static} group takes 53 first-person and 28 third-person WBench~\cite{wbench} cases whose obstacle can be reached within the clip; the \emph{dynamic} group has 50 first frames in WBench's style, 30 first-person and 20 third-person, with moving people, crowds, vehicles, animals and objects. On the WBench cases the input is ours, because WBench's own navigation turns rarely reach an obstacle: an 8\,s input that drives into the obstacle and always moves forward. There is no 3D ground truth here, so four VLM judges apply the rule of Task~2 to what can be seen, and Task~3 reports the share of videos with a violation. A Task~3 video is \emph{safe and with the input} when the judge finds neither a violation nor going back; the main table's Overall score is the mean of SIS and this share of Task~3 videos.

\begin{table*}[t]
\centering\footnotesize
\caption{\textbf{Six generators and their heads.} Architecture values are as configured in the checkpoints we run. Collision head: first denoising forward of the current chunk; prediction head: commit pass of the previous chunk. Operating point: temperature $T$, safety threshold $\tau$ and keep margin $\delta$ of \cref{eq:select}, selected on validation videos. All heads use the same 729 training videos and run for 1,000 optimizer steps (\cref{s:train}).}
\label{s:tab-models}
\setlength{\tabcolsep}{4pt}
\resizebox{\textwidth}{!}{%
\begin{tabular}{@{}lcccccc@{}}
\toprule
 & LingBot-World 2.0~\cite{lingbotv2} & HY-WorldPlay~\cite{worldplay} & Echo-WM~\cite{echowm,echo15} & Evoke-Turbo~\cite{evoke} & WorldPlay2~\cite{worldplay2} & Hy-Nomas~\cite{hynomas}\\
\midrule
DiT blocks $\times$ width $D$ & $40\times5120$ & $54\times2048$ & $48\times4096$ & $40\times5120$ & $40\times5120$ (2 experts) & $54\times2048$ (dual-stream)\\
Video, fps & $464\times832$, 16 & $480\times832$, 24 & $704\times1280$, 24 & $384\times640$, 24 & $448\times832$, 16 & $704\times1280$, 24\\
Latents per chunk $F$ & 3 (0.75\,s) & 4 (0.67\,s) & 3 (1\,s) & 9 (1.5\,s) & 4 (1\,s) & 4 (0.67\,s)\\
Tokens per latent $R\times C$ & $29\times52$ & $30\times52$ & $22\times40$ & $6\times10$ at step 0 & $28\times52$ & $22\times40$\\
Control signal & pose $\to$ Pl\"ucker, per block & pose attention + action label & pose $\to$ attention branch & pose $\to$ point-cloud warp & per-latent action row & relative pose $\to$ cross-attention, per block\\
Head's camera input & own encoder, per token & rays + frozen label & Pl\"ucker ray per token & Pl\"ucker ray per token & own encoder, per latent & rays + frozen camera token\\
\midrule
Collision head: blocks & 30, 38, 39 & 27 & 42 & 20 & 30 & 33\\
Prediction head: blocks & 30 & 33 & 30 & \na & 20 & 33\\
Parameters (collision / prediction) & 1.72M / 1.74M & 0.95M / 0.97M & 0.95M / 0.97M & 1.07M / \na & 1.72M / 1.74M & 0.70M / 0.71M\\
Operating point $T$, $\tau$, $\delta$ (collision) & .82, .85, .030 & .743, .80, .150 & .906, .85, .000 & .82, .80, .100 & .82, .95, .030 & .906, .85, .000\\
Operating point $T$, $\tau$, $\delta$ (prediction) & .906, .85, .010 & .82, .80, .000 & 1.00, .90, .020 & \na & .82, .95, .075 & .906, .85, .000\\
\bottomrule
\end{tabular}}
\end{table*}
\subsection{The evaluator}
\label{s:bench-eval}
\label{s:rules}
\label{s:evaluators}
Task~2 is judged by geometry, conservatively: a violation is called only when it exceeds what the measurement could get wrong. The generated frames are put after the true history. VGGT-$\Omega$~\cite{vggtomega} reconstructs depth and camera poses of the last 2\,s of history at 4\,fps and of the 5\,s after the hand-off at 8\,fps (49 frames) in one multi-view pass; one scale, the true length of the sampled history path over its estimated length, is applied to every position and depth. The estimated path is placed through the true hand-off pose on the case region's occupancy map at camera height (5\,cm cells), whose free space is the set of cells connected to the hand-off camera. SAM~2~\cite{sam2} tracks the obstacle from the true hand-off mask, only to detect a vanishing obstacle. Each sample also gives the depth $D_t$ in the image window around the commanded direction (20th percentile, middle third of the width). \Cref{s:tab-verdicts} gives the rule in its order with the origin of every threshold; ``at the obstacle'' means within 0.6\,m of the obstacle's points, or $D_t\le0.6$\,m.

The \emph{Succ.} column in \cref{s:tab-l2-full} measures safe arrival. It counts routes that reach the obstacle without a violation or retreat (\emph{stopped}, \emph{slid}, or \emph{stopped, not slid}). Thus a route can count toward \emph{Succ.} even when its slide falls short of the reference. An \emph{early stop} also counts if the closest approach is within one chunk of travel, computed from the generator's chunk duration and the commanded speed.

\subsection{How far the evaluators can be trusted}
\label{s:bench-trust}

\paragraph{Unreal renders of known behaviour.}
On Unreal renders whose poses are known, the evaluator is right on 366 of 391 cases (199 of 219 reference executions that collide and slide, 149 of 152 walk-throughs, and 18 of 20 renders of other behaviours). Over 3,350 generated Task~2 videos, VGGT-$\Omega$ and an independent reconstruction, DA3~\cite{da3}, agree on the violation verdict with $\kappa=0.89$.

\paragraph{The Task~3 judges against people.}
\label{s:l3-judging}
Four VLMs judge every Task~3 video under one fixed prompt, tuned on Task~2 development videos only: two closed models, Claude Opus~5.5 and GPT-6 Astra, and two open ones, DeepSeek V4.1 Flash and Qwen3.8-27B, for open-weight reference. Three annotators who are not authors of this paper label a sample drawn at random with a fixed seed (56 Task~2 videos mixed with 224 Task~3 videos) under a fixed written guideline (pairwise $\kappa$ on violation 0.78--0.80). Of the 280 sampled videos, 269 have a majority violation or no-violation label (44 violations); the other 11 lack two matching binary labels because annotators could mark a video as unclear or corrupted. Recall and precision of violation are 0.91 and 0.78 for Claude, 0.81 and 0.92 for GPT-6, 0.59 and 0.70 for Qwen, and 0.55 and 0.71 for DeepSeek, and on generated Task~2 videos the judges agree with the geometric evaluator with $\kappa$ of 0.81 (GPT-6 and Claude), 0.68 (Qwen) and 0.67 (DeepSeek). The open judges agree less with people and with geometry, so the main table reports the mean of Claude and GPT-6.

\section{Method}
\label{s:method}

\subsection{The six generators}
\label{s:gen}

Across all six generators, the head design is shared. The tapped blocks, token projection ($D\to128$) and candidate encoder follow each generator's architecture (\cref{s:tab-models}).

A candidate is one of the 34 ground displacements of \cref{s:cands}, expressed over the model's own chunk duration at the commanded speed and passed through the model's own control path, so the head scores exactly the input the generator would receive. Evoke has no prediction head because it keeps no memory forward between chunks.

\subsection{The head}
\label{s:head}
\label{s:problem}
The prediction head reads the commit-pass tokens $H_{k-1}^{\text{com}}$ of \cref{eq:commit}, the collision head the step-0 tokens $H_k^{\text{step}}$, both through \cref{eq:head}. The token projector $\psi_H$ passes every token through its own learned linear map $W_s\in\mathbb{R}^{128\times D}$ and a LayerNorm. Nothing is pooled and no fixed projection is used: on LingBot the prediction head attends over all $1\cdot3\cdot29\cdot52=4{,}524$ tokens of block 30 and the collision head over 13,572 tokens of blocks 30, 38 and 39. The frozen control encoder $E$ gives one $D$-dimensional vector per token, which $\psi_c$ maps through $W_c$ and adds to the state token at the same (frame, row, column). The learned query $q$ reads the resulting $\tilde H_i$ through two cross-attention blocks (4 heads, width 128) and outputs the safety logit $\ell_i$, the swept clearance $\hat g_i$ and the intent score $\hat A_i$. Written out, \cref{eq:head} is
\begingroup\allowdisplaybreaks[4]\begin{align*}
\psi_H(H) &= \mathrm{LN}(W_s H) + e_{\text{block}} + e_{\text{frame}} + e_{\text{row}} + e_{\text{col}},\\
\psi_c(\cdot) &= \mathrm{LN}(W_c E(c_i)) + \mathrm{mix}\big(\mathrm{LN}(W_c E(u_k))\big) + e_{\text{time}},\\
\tilde H_i &= \psi_H(H) + \psi_c\big(E(c_i),E(u_k)\big),
\end{align*}\endgroup
with $W_s,W_c\in\mathbb{R}^{128\times D}$ applied per token. The mix term adds the input $u_k$, which arrived after $H_{k-1}^{\text{com}}$ was computed; the collision head reads $H_k^{\text{step}}$, which already holds it, without that term.

\paragraph{Selection.}
\Cref{eq:select} is applied with $\Phat_i=\sigma(\ell_i/T)$ over the feasible set $\mathcal{S}_k=\{i:\Phat_i\ge\tau\}$, and $a_k^\star$ is a stop if $\mathcal{S}_k$ is empty. \Cref{s:tab-models} gives every model's $(T,\tau,\delta)$.

\subsection{Candidates}
\label{s:cands}
Each decision scores 34 ground displacements $c_i\in\mathcal{C}$: the input $c_0=u_k$, a stop, the input shortened to 75, 50, 25 and 12.5\%, half-length turns at 30\textdegree{} and 60\textdegree{} to each side, and 12 per side on an intent ellipse. The ellipse points sit at deviations $\alpha$ of 8.8\textdegree{} to 88\textdegree{} from the input with length $\rho(\alpha)=\min\big(1,\cos\alpha/(\cos^2\alpha+w\sin^2\alpha)\big)$, $w=0.5$: the length in direction $\alpha$ that minimises the intent error when its lateral part is weighted by $w$, yielding shorter steps for larger turns.

\subsection{Labels}
\label{s:labels}
Labels come from the occupancy map of the Unreal level (\cref{eq:labels}): a disc of radius 0.25\,m sweeps along each candidate $c_i$ at camera height and gives the clearance $g_i$, clipped to $[-0.25,1]$\,m, and the soft safety label $P_i$ with margin $m=0.05$\,m and scale 0.03\,m. The intent score $A_i$ compares the candidate's displacement $d_i$ with the collide-and-slide displacement $v_S(u_k)$ of the input through the error $\epsilon_i^2=\big(((d_i-v_S)\cdot \hat t)^2+w\,((d_i-v_S)\cdot \hat t_\perp)^2\big)/|u_k|^2$, with $\hat t$ the slide direction; at a head-on wall the stop has $A_i=1$.

\subsection{Training and validation}
\label{s:train}\label{s:select}
Each head trains online while the frozen generator replays the training videos with their true latents and poses, updating on the tapped tokens of each decision state as they are computed. The loss is \cref{eq:loss}, where the Huber term switches from quadratic to linear at 0.05\,m. All heads across the six generators are trained for 1,000 optimizer steps on the same 729 UnrealZoo videos from 44 scenes, rendered at each model's native frame rate (\cref{s:tab-models}).

Readout blocks and operating points $(T,\tau,\delta)$ are selected on held-out validation videos from scenes disjoint from both the training and benchmark splits. For each head, temperature $T$ is fitted by negative log-likelihood on the soft safety labels, and $(\tau,\delta)$ are chosen on a grid to minimize validation regret under a bounded missed-intervention rate. All deployed block indices and operating points are listed in \cref{s:tab-models}, and all trained head weights (both collision and prediction modes) are publicly released.

\begin{table}[!t]
\centering\footnotesize
\caption{\textbf{Task~1 on every decision of each mode} (100 routes). The main paper's \cref{tab:decisions} scores both modes on the decisions both make (last column); here each mode is scored on all of its own. Collision mode also decides at the first replayed chunk of a route where prediction mode has no prior commit pass (all dual-mode models except Hy-Nomas, which runs a commit pass on the history clip). Missed, False and Score as in the main paper.}
\label{s:tab-l1-own}
\setlength{\tabcolsep}{3.5pt}
\resizebox{\columnwidth}{!}{%
\begin{tabular}{@{}llccccc@{}}
\toprule
 & Mode & Decisions & Missed$\downarrow$ & False$\downarrow$ & Score$\uparrow$ & Common\\
\midrule
LingBot-World 2.0 & Collision & 1{,}107 & 2.6 & 6.1 & 90.2 & 1{,}007\\
 & Prediction & 1{,}007 & 2.4 & 9.3 & 88.1 & 1{,}007\\
HY-WorldPlay & Collision & 1{,}238 & 4.0 & 8.2 & 83.2 & 1{,}138\\
 & Prediction & 1{,}138 & 6.6 & 29.9 & 79.7 & 1{,}138\\
Echo-WM & Collision & 801 & 3.3 & 21.2 & 84.4 & 701\\
 & Prediction & 701 & 2.5 & 12.7 & 84.6 & 701\\
Evoke-Turbo & Collision & 408 & 11.5 & 4.4 & 79.7 & 408\\
WorldPlay2 & Collision & 823 & 0.6 & 12.7 & 87.2 & 723\\
 & Prediction & 723 & 1.4 & 17.9 & 82.6 & 723\\
Hy-Nomas & Collision & 1{,}117 & 1.5 & 49.4 & 76.7 & 1{,}117\\
 & Prediction & 1{,}117 & 5.7 & 31.1 & 72.7 & 1{,}117\\
\bottomrule
\end{tabular}}
\end{table}

\subsection{Baselines: specification}
\label{s:baselines}
\paragraph{Text prompt.} One sentence is appended to the VLM scene caption each raw run uses:
\begin{quote}\small\itshape
There is \textup{$\langle$obstacle$\rangle$} \textup{$\langle$side$\rangle$}, about \textup{$\langle d\rangle$}\,m away. The camera must not go through the \textup{$\langle$obstacle$\rangle$}: on reaching it, the camera must stop or slide along it.
\end{quote}
The three fields use ground truth that the head never receives, so the prompt carries more information about the obstacle than our method does. On Task~2, $\langle$obstacle$\rangle$ denotes the target obstacle in the hand-off frame and $\langle d\rangle$ is the median ground-truth depth within its mask, rounded to 0.5\,m.
On Task~3, where metric depth is unavailable, $\langle$obstacle$\rangle$, $\langle$side$\rangle$, and approximate distance $\langle d\rangle$ are annotated from the starting image. No negative prompt is used, since the six samplers run without classifier-free guidance. The open-path routes have no obstacle for the prompt to name, so the text prompt does not apply there.

\paragraph{External depth (stop).} Before each decision, DA3~\cite{da3} reconstructs the true history together with every frame decoded since the hand-off, with scale fitted on the true camera path. The input's swept clearance on this map gives a safety probability by \cref{eq:labels}; the guard keeps the input if it meets a safety threshold selected on the validation set and stops otherwise. On Task~3, with no true history, scale and level come from the floor at an assumed eye height of 1.6\,m.

\section{Results}
\label{s:results}

Verdicts on generated video in this section follow the evaluator's rule (\cref{s:rules}). Unless a caption says otherwise, intervals are 95\% from 4000 case resamples (seed 0), and differences are paired by case; scene resamples, where a caption names them, draw the cases of one scene together.

\subsection{Task~1 in full}
\label{s:res-l1}

Task~1 scores each head's decisions on the true history of the 100 routes, without generation (\cref{s:bench-l1}). The main paper's Task~1 table scores both modes on the decisions both make; \cref{s:tab-l1-own} gives each mode on all of its own decisions.

\clearpage
\begin{table}[!t]
\centering\footnotesize
\caption{\textbf{Task~2 in full} (percent, 100 routes; VGGT-$\Omega$, one evaluator rule). Viol.: any violation. Back: going back against the input. SFP: mean progress towards the end point of the engine's own execution, zero with a violation (\cref{s:l2-full}). Slide: the share of the engine's slide the video makes, on the 65 routes where the engine slides at least 0.5\,m. Open: SFP on the 50 open-path routes (n/a: no obstacle for the text prompt to name). Succ.: the camera reaches the obstacle without a violation or going back and then stops or slides, a stop within one chunk of travel included (\cref{s:rules}). SIS: the main paper's score, with its 95\% interval. A camera that stops exactly at the contact on every route would have SFP 72 and Slide 0.}
\label{s:tab-l2-full}
\setlength{\tabcolsep}{3.5pt}
\resizebox{\columnwidth}{!}{%
\renewcommand{\arraystretch}{1.15}
\begin{tabular}{@{}lccccccc@{}}
\toprule
Method & Viol.$\downarrow$ & Back$\downarrow$ & SFP$\uparrow$ & Slide$\uparrow$ & Open$\uparrow$ & Succ.$\uparrow$ & SIS$\uparrow$\\
\midrule
\multicolumn{8}{@{}l}{\textbf{LingBot-World 2.0}}\\
\hspace{5pt}Raw model & 89 & 5 & 6 & 4 & 94 & 6 & 5\,{\scriptsize[1, 9]}\\
\hspace{5pt}Text prompt & 94 & 2 & 2 & 3 & n/a & 4 & 4\,{\scriptsize[1, 8]}\\
\hspace{5pt}External depth (stop) & 2 & 0 & 55 & 2 & 92 & 72 & 34\,{\scriptsize[28, 41]}\\
\hspace{5pt}NoPeWorld, collision mode & 7 & 0 & 61 & 19 & 93 & 87 & 50\,{\scriptsize[42, 58]}\\
\hspace{5pt}NoPeWorld, prediction mode & 4 & 0 & 58 & 30 & 92 & 87 & 56\,{\scriptsize[48, 63]}\\
\midrule
\multicolumn{8}{@{}l}{\textbf{HY-WorldPlay}}\\
\hspace{5pt}Raw model & 86 & 1 & 8 & 2 & 93 & 12 & 5\,{\scriptsize[2, 10]}\\
\hspace{5pt}Text prompt & 83 & 2 & 10 & 5 & n/a & 14 & 8\,{\scriptsize[4, 14]}\\
\hspace{5pt}External depth (stop) & 0 & 0 & 59 & 4 & 92 & 82 & 38\,{\scriptsize[30, 46]}\\
\hspace{5pt}NoPeWorld, collision mode & 15 & 2 & 42 & 14 & 91 & 67 & 38\,{\scriptsize[32, 44]}\\
\hspace{5pt}NoPeWorld, prediction mode & 21 & 1 & 49 & 23 & 92 & 70 & 43\,{\scriptsize[36, 52]}\\
\midrule
\multicolumn{8}{@{}l}{\textbf{Echo-WM}}\\
\hspace{5pt}Raw model & 86 & 1 & 7 & 7 & 91 & 12 & 11\,{\scriptsize[6, 17]}\\
\hspace{5pt}Text prompt & 86 & 0 & 7 & 10 & n/a & 13 & 12\,{\scriptsize[7, 18]}\\
\hspace{5pt}External depth (stop) & 2 & 3 & 32 & 2 & 89 & 40 & 14\,{\scriptsize[9, 19]}\\
\hspace{5pt}NoPeWorld, collision mode & 3 & 6 & 52 & 18 & 87 & 73 & 44\,{\scriptsize[36, 52]}\\
\hspace{5pt}NoPeWorld, prediction mode & 5 & 2 & 58 & 12 & 88 & 78 & 42\,{\scriptsize[34, 50]}\\
\midrule
\multicolumn{8}{@{}l}{\textbf{Evoke-Turbo}}\\
\hspace{5pt}Raw model & 77 & 14 & 11 & 5 & 94 & 9 & 8\,{\scriptsize[3, 14]}\\
\hspace{5pt}Text prompt & 80 & 9 & 10 & 7 & n/a & 10 & 10\,{\scriptsize[4, 16]}\\
\hspace{5pt}External depth (stop) & 4 & 5 & 24 & 1 & 92 & 33 & 11\,{\scriptsize[7, 15]}\\
\hspace{5pt}NoPeWorld, collision mode & 30 & 11 & 37 & 13 & 93 & 40 & 27\,{\scriptsize[18, 35]}\\
\midrule
\multicolumn{8}{@{}l}{\textbf{WorldPlay2}}\\
\hspace{5pt}Raw model & 11 & 36 & 25 & 31 & 90 & 48 & 36\,{\scriptsize[29, 43]}\\
\hspace{5pt}Text prompt & 9 & 30 & 34 & 36 & n/a & 51 & 43\,{\scriptsize[36, 50]}\\
\hspace{5pt}External depth (stop) & 1 & 20 & 44 & 5 & 89 & 69 & 31\,{\scriptsize[24, 39]}\\
\hspace{5pt}NoPeWorld, collision mode & 1 & 30 & 43 & 6 & 88 & 62 & 25\,{\scriptsize[18, 31]}\\
\hspace{5pt}NoPeWorld, prediction mode & 4 & 24 & 40 & 8 & 89 & 58 & 29\,{\scriptsize[23, 35]}\\
\midrule
\multicolumn{8}{@{}l}{\textbf{Hy-Nomas}}\\
\hspace{5pt}Raw model & 95 & 2 & 2 & 0 & 6 & 2 & 2\,{\scriptsize[0, 5]}\\
\hspace{5pt}Text prompt & 95 & 0 & 2 & 2 & n/a & 5 & 5\,{\scriptsize[1, 9]}\\
\hspace{5pt}External depth (stop) & 50 & 16 & 23 & 11 & 18 & 27 & 20\,{\scriptsize[12, 28]}\\
\hspace{5pt}NoPeWorld, collision mode & 40 & 11 & 24 & 17 & 20 & 42 & 31\,{\scriptsize[22, 41]}\\
\hspace{5pt}NoPeWorld, prediction mode & 45 & 14 & 20 & 24 & 19 & 34 & 29\,{\scriptsize[19, 39]}\\
\bottomrule
\end{tabular}}
\end{table}

\subsection{Task~2 in full}
\label{s:l2-full}
This subsection gives Task~2, closed-loop generation on the 100 routes and on the 50 open-path routes, in full. The main table reports SIS, violation rate and Slide on the 100 obstacle routes. \Cref{s:tab-l2-full} gives every score we computed. SFP, safe final progress, is the mean over routes of $\mathrm{clip}\big(1-\max(0,d-\varepsilon)/(D-\varepsilon),0,1\big)$, with $d$ the distance from the camera's last position to the end point of the engine's own execution, $D$ that end point's distance from the hand-off and $\varepsilon=0.3$\,m, and zero for a video with a violation. Slide is the share of the engine's slide that the video makes, on the 65 routes where the engine slides at least 0.5\,m, and is zero for rollouts that violate or retreat.

\newpage
\begin{table}[!t]
\centering\footnotesize
\caption{\textbf{Routes with an open path}: percent of the 50 unobstructed control routes on which a method stops early (in brackets: goes back). On Hy-Nomas, violation rates are 22\% for Raw, 4\% for Depth, 10\% for Coll and 12\% for Pred. \na: mode unavailable.}
\label{s:tab-l2-ctl}
\setlength{\tabcolsep}{3pt}
\resizebox{\columnwidth}{!}{%
\renewcommand{\arraystretch}{1.45}
\begin{tabular}{@{}l cccc@{}}
\toprule
 & Raw & Depth (stop) & NoPeWorld, coll. & NoPeWorld, pred.\\
\midrule
LingBot-World 2.0 & 0 & 4 & 2 & 0\\
HY-WorldPlay & 0 & 2 & 6 & 4\\
Echo-WM & 0 & 6 & 6 & 6\\
Evoke-Turbo & 0 & 2 & 2 & \na\\
WorldPlay2 & 4 & 2 & 4 (4) & 6\\
Hy-Nomas & 0 (6) & 0 (8) & 0 & 0 (8)\\
\bottomrule
\end{tabular}}
\end{table}

\begin{table}[!t]
\centering\footnotesize
\caption{\textbf{How far ahead the head acts}: seconds from the first change of the input (Task~2) or the first flag of the input as unsafe (Task~1) to the commanded path's first contact in Unreal geometry, median \interval{25th}{75th percentile}. $\rho$: Spearman correlation between time-to-contact from hand-off and first-change time on the 51 routes with contact 0.5--3\,s after hand-off.}
\label{s:tab-leadtime}
\setlength{\tabcolsep}{3pt}\renewcommand{\arraystretch}{1.05}
\resizebox{\columnwidth}{!}{%
\renewcommand{\arraystretch}{1.40}
\begin{tabular}{@{}ll c cc cc@{}}
\toprule
 & & Task~2 (100) & \multicolumn{2}{c}{Contact 0.5--3\,s (51)} & \multicolumn{2}{c}{Task~1 (100)}\\
\cmidrule(lr){3-3}\cmidrule(lr){4-5}\cmidrule(l){6-7}
Model & Mode & Lead (s) & Lead (s) & $\rho$ & Flag (s) & Unsafe (s)\\
\midrule
LingBot-World 2.0 & Coll & 0.31 \interval{0.25}{0.56} & 0.44 \interval{0.19}{0.56} & 0.88 & 0.50 \interval{0.30}{0.75} & 0.44\\
 & Pred & 0.31 \interval{0.25}{0.62} & 0.50 \interval{0.25}{0.62} & 0.92 & 0.56 \interval{0.28}{0.78} & 0.44\\
\midrule
HY-WorldPlay & Coll & 0.62 \interval{0.33}{1.00} & 0.69 \interval{0.36}{0.90} & 0.84 & 0.56 \interval{0.38}{0.81} & 0.45\\
 & Pred & 0.91 \interval{0.40}{1.06} & 0.88 \interval{0.56}{1.21} & 0.83 & 0.60 \interval{0.39}{0.79} & 0.45\\
\midrule
Echo-WM & Coll & 1.00 \interval{0.52}{1.06} & 0.75 \interval{0.50}{1.25} & 0.73 & 0.75 \interval{0.44}{1.06} & 0.56\\
 & Pred & 0.88 \interval{0.44}{1.06} & 0.69 \interval{0.44}{0.94} & 0.69 & 0.81 \interval{0.50}{1.12} & 0.56\\
\midrule
Evoke-Turbo & Coll & 1.00 \interval{0.75}{1.06} & 0.75 \interval{0.56}{1.03} & 0.76 & 0.73 \interval{0.35}{1.10} & 0.82\\
\midrule
WorldPlay2 & Coll & 1.00 \interval{0.62}{1.06} & 0.69 \interval{0.50}{0.94} & 0.84 & 0.88 \interval{0.50}{1.19} & 0.56\\
 & Pred & 1.00 \interval{0.72}{1.06} & 0.84 \interval{0.56}{1.11} & 0.77 & 0.94 \interval{0.56}{1.38} & 0.56\\
\midrule
Hy-Nomas & Coll & 1.00 \interval{0.88}{1.19} & 1.19 \interval{0.77}{1.62} & 0.43 & 0.72 \interval{0.44}{1.02} & 0.44\\
 & Pred & 0.84 \interval{0.40}{1.06} & 0.76 \interval{0.54}{1.21} & 0.70 & 0.65 \interval{0.44}{1.00} & 0.44\\
\bottomrule
\end{tabular}}
\end{table}

\subsubsection{Routes with an open path}
\label{s:l2-ctl}
On 50 further routes nothing blocks the commanded path for its whole length, testing whether a method halts unnecessarily when the way ahead is clear (\cref{s:tab-l2-ctl}).

\subsection{When the head acts}
\label{s:leadtime}
\label{s:behaviour}
\Cref{s:tab-leadtime} measures how long before contact the head intervenes: in Task~2, the time from the start of the first chunk whose input it changes to the commanded path's first contact with the obstacle in Unreal geometry; in Task~1, on true history, when it first flags keeping the input as unsafe alongside when the ground-truth labels first mark it unsafe. Because lead time is quantized in steps of one chunk and bounded by the interval from hand-off to contact, we also report it separately on the 51 routes where contact occurs 0.5--3\,s after hand-off. For every generator and mode, a later time to contact is associated with a later first intervention ($\rho$ in \cref{s:tab-leadtime}).

\onecolumn
\begin{table}[!t]
\centering\small
\caption{\textbf{How much of the input is kept} (Task~2, \%) where Unreal stops (35 routes), slides (65) or the path is open (50). Gap: median closest distance to the obstacle over the videos without violation or going back (their number). Mean: generators with that run. n/a: an open path has no obstacle for the text prompt to name.}
\label{s:tab-intent}
\setlength{\tabcolsep}{4pt}
\renewcommand{\arraystretch}{1.04}\begin{tabular}{@{}l ccc ccc ccc@{}}
\toprule
 & \multicolumn{3}{c}{Unreal stops (35 routes)} & \multicolumn{3}{c}{Unreal slides (65 routes)} & \multicolumn{3}{c}{Open path (50 routes)}\\
\cmidrule(lr){2-4}\cmidrule(lr){5-7}\cmidrule(lr){8-10}
Method & Viol.$\downarrow$ & Reached$\uparrow$ & Gap (m) & Viol.$\downarrow$ & Slide$\uparrow$ & Slid $\ge$0.5\,m$\uparrow$ & Progress$\uparrow$ & Changed$\downarrow$ & Stop/back$\downarrow$\\
\midrule
\multicolumn{10}{@{}l}{\textbf{LingBot-World 2.0}}\\
\hspace{5pt}Raw model & 89 & 6 & 0.08\,{\scriptsize(2)} & 89 & 4 & 3 & 90 & 0 & 0\\
\hspace{5pt}Text prompt & 91 & 6 & 0.05\,{\scriptsize(2)} & 95 & 3 & 3 & n/a & n/a & n/a\\
\hspace{5pt}External depth & 0 & 43 & 0.68\,{\scriptsize(35)} & 3 & 2 & 5 & 88 & 6 & 4\\
\hspace{5pt}Ours (Coll) & 3 & 83 & 0.54\,{\scriptsize(34)} & 9 & 19 & 15 & 89 & 4 & 2\\
\hspace{5pt}Ours (Pred) & 3 & 80 & 0.49\,{\scriptsize(34)} & 5 & 30 & 31 & 88 & 8 & 0\\
\midrule
\multicolumn{10}{@{}l}{\textbf{HY-WorldPlay}}\\
\hspace{5pt}Raw model & 86 & 9 & 0.26\,{\scriptsize(4)} & 86 & 2 & 0 & 94 & 0 & 0\\
\hspace{5pt}Text prompt & 83 & 11 & 0.11\,{\scriptsize(5)} & 83 & 5 & 6 & n/a & n/a & n/a\\
\hspace{5pt}External depth & 0 & 54 & 0.59\,{\scriptsize(35)} & 0 & 4 & 5 & 92 & 4 & 2\\
\hspace{5pt}Ours (Coll) & 17 & 57 & 0.43\,{\scriptsize(28)} & 14 & 14 & 18 & 92 & 12 & 6\\
\hspace{5pt}Ours (Pred) & 29 & 63 & 0.35\,{\scriptsize(25)} & 17 & 23 & 26 & 93 & 32 & 4\\
\midrule
\multicolumn{10}{@{}l}{\textbf{Echo-WM}}\\
\hspace{5pt}Raw model & 77 & 17 & 0.41\,{\scriptsize(7)} & 91 & 7 & 6 & 100 & 0 & 0\\
\hspace{5pt}Text prompt & 83 & 14 & 0.22\,{\scriptsize(6)} & 88 & 10 & 8 & n/a & n/a & n/a\\
\hspace{5pt}External depth & 3 & 9 & 1.12\,{\scriptsize(34)} & 2 & 2 & 5 & 97 & 8 & 6\\
\hspace{5pt}Ours (Coll) & 3 & 60 & 0.58\,{\scriptsize(34)} & 3 & 18 & 18 & 96 & 28 & 6\\
\hspace{5pt}Ours (Pred) & 6 & 63 & 0.59\,{\scriptsize(33)} & 5 & 12 & 12 & 96 & 12 & 6\\
\midrule
\multicolumn{10}{@{}l}{\textbf{Evoke-Turbo}}\\
\hspace{5pt}Raw model & 80 & 11 & 0.16\,{\scriptsize(4)} & 75 & 5 & 6 & 95 & 0 & 0\\
\hspace{5pt}Text prompt & 80 & 14 & 0.14\,{\scriptsize(5)} & 80 & 7 & 8 & n/a & n/a & n/a\\
\hspace{5pt}External depth & 3 & 9 & 1.33\,{\scriptsize(33)} & 5 & 1 & 3 & 94 & 6 & 2\\
\hspace{5pt}Ours (Coll) & 40 & 29 & 0.67\,{\scriptsize(17)} & 25 & 13 & 14 & 94 & 6 & 2\\
\midrule
\multicolumn{10}{@{}l}{\textbf{WorldPlay2}}\\
\hspace{5pt}Raw model & 17 & 37 & 0.57\,{\scriptsize(16)} & 8 & 31 & 32 & 94 & 0 & 4\\
\hspace{5pt}Text prompt & 9 & 46 & 0.50\,{\scriptsize(20)} & 9 & 36 & 40 & n/a & n/a & n/a\\
\hspace{5pt}External depth & 0 & 23 & 0.74\,{\scriptsize(29)} & 2 & 5 & 5 & 93 & 8 & 2\\
\hspace{5pt}Ours (Coll) & 0 & 29 & 0.67\,{\scriptsize(20)} & 2 & 6 & 5 & 91 & 14 & 8\\
\hspace{5pt}Ours (Pred) & 6 & 34 & 0.71\,{\scriptsize(22)} & 3 & 8 & 9 & 92 & 14 & 6\\
\midrule
\multicolumn{10}{@{}l}{\textbf{Hy-Nomas}}\\
\hspace{5pt}Raw model & 91 & 6 & 0.17\,{\scriptsize(3)} & 97 & 0 & 0 & 99 & 0 & 6\\
\hspace{5pt}Text prompt & 89 & 11 & 0.14\,{\scriptsize(4)} & 98 & 2 & 2 & n/a & n/a & n/a\\
\hspace{5pt}External depth & 51 & 26 & 0.41\,{\scriptsize(12)} & 49 & 11 & 9 & 99 & 30 & 8\\
\hspace{5pt}Ours (Coll) & 43 & 49 & 0.34\,{\scriptsize(18)} & 38 & 17 & 20 & 98 & 92 & 0\\
\hspace{5pt}Ours (Pred) & 51 & 34 & 0.20\,{\scriptsize(13)} & 42 & 24 & 22 & 99 & 72 & 8\\
\midrule
\multicolumn{10}{@{}l}{\textbf{Mean over generators}}\\
\hspace{5pt}Raw model & 73 & 14 & \na & 74 & 8 & 8 & 95 & 0 & 2\\
\hspace{5pt}Text prompt & 72 & 17 & \na & 76 & 10 & 11 & n/a & n/a & n/a\\
\hspace{5pt}External depth & 10 & 27 & \na & 10 & 4 & 5 & 94 & 10 & 4\\
\hspace{5pt}Ours (Coll) & 18 & 51 & \na & 15 & 15 & 15 & 94 & 26 & 4\\
\hspace{5pt}Ours (Pred) & 19 & 55 & \na & 14 & 19 & 20 & 94 & 28 & 5\\
\bottomrule
\end{tabular}
\end{table}

\subsection*{How much of the Task~2 input is kept}
\Cref{s:tab-intent} groups routes by the engine's response to the command. Reached counts videos that reach the obstacle without a violation or retreat. On open paths, Progress is displacement along the command divided by its length, capped at one. Changed counts routes with at least one input replacement. Stop/back counts early stops, nonresponses and retreats.

\clearpage
\begin{table}[!t]
\centering\small
\caption{\textbf{Task~3 by judge}: violation rate (\%) of each generator and method for each of the four VLM judges. Static: WBench, first and third person (53 + 28 videos); dynamic: moving obstacles, first and third person (30 + 20); parts pooled by size. The main table reports the mean of Claude and GPT-6. Qwen's rates are over its valid answers (99.3\% of the videos).}
\label{s:tab-l3-compact}
\setlength{\tabcolsep}{4pt}
\renewcommand{\arraystretch}{1.00}\begin{tabular}{@{}l cc cc cc cc@{}}
\toprule
 & \multicolumn{2}{c}{Claude} & \multicolumn{2}{c}{GPT-6} & \multicolumn{2}{c}{DeepSeek} & \multicolumn{2}{c}{Qwen}\\
\cmidrule(lr){2-3}\cmidrule(lr){4-5}\cmidrule(lr){6-7}\cmidrule(lr){8-9}
Method & Static & Dynamic & Static & Dynamic & Static & Dynamic & Static & Dynamic\\
\midrule
\multicolumn{9}{@{}l}{\textit{LingBot-World 2.0}~\cite{lingbotv2}}\\
\hspace{4pt}Raw model & 72 & 22 & 63 & 14 & 46 & 22 & 41 & 34\\
\hspace{4pt}Text prompt & 68 & 32 & 72 & 12 & 43 & 24 & 41 & 35\\
\hspace{4pt}External depth & 31 & 14 & 30 & 8 & 17 & 18 & 20 & 31\\
\rowcolor{ours}\hspace{4pt}Ours (Coll) & 20 & 16 & 19 & 0 & 7 & 12 & 6 & 23\\
\rowcolor{ours}\hspace{4pt}Ours (Pred) & 21 & 14 & 17 & 8 & 5 & 22 & 9 & 36\\
\specialrule{\lightrulewidth}{1pt}{1pt}
\multicolumn{9}{@{}l}{\textit{HY-WorldPlay}~\cite{worldplay}}\\
\hspace{4pt}Raw model & 25 & 14 & 20 & 8 & 16 & 12 & 15 & 22\\
\hspace{4pt}Text prompt & 27 & 12 & 21 & 4 & 14 & 8 & 16 & 24\\
\hspace{4pt}External depth & 17 & 4 & 14 & 0 & 5 & 8 & 6 & 14\\
\rowcolor{ours}\hspace{4pt}Ours (Coll) & 19 & 0 & 14 & 0 & 9 & 2 & 7 & 10\\
\rowcolor{ours}\hspace{4pt}Ours (Pred) & 14 & 0 & 10 & 0 & 6 & 4 & 6 & 14\\
\specialrule{\lightrulewidth}{1pt}{1pt}
\multicolumn{9}{@{}l}{\textit{Echo-WM}~\cite{echowm}}\\
\hspace{4pt}Raw model & 64 & 32 & 57 & 12 & 42 & 28 & 35 & 34\\
\hspace{4pt}Text prompt & 62 & 32 & 58 & 22 & 42 & 36 & 38 & 40\\
\hspace{4pt}External depth & 37 & 16 & 30 & 6 & 16 & 12 & 17 & 24\\
\rowcolor{ours}\hspace{4pt}Ours (Coll) & 25 & 14 & 19 & 2 & 5 & 12 & 6 & 20\\
\rowcolor{ours}\hspace{4pt}Ours (Pred) & 15 & 10 & 15 & 4 & 4 & 14 & 5 & 20\\
\specialrule{\lightrulewidth}{1pt}{1pt}
\multicolumn{9}{@{}l}{\textit{Evoke-Turbo}~\cite{evoke}}\\
\hspace{4pt}Raw model & 49 & 28 & 46 & 16 & 22 & 26 & 19 & 35\\
\hspace{4pt}Text prompt & 51 & 32 & 49 & 14 & 28 & 32 & 21 & 36\\
\hspace{4pt}External depth & 27 & 10 & 30 & 4 & 14 & 10 & 14 & 16\\
\rowcolor{ours}\hspace{4pt}Ours (Coll) & 17 & 2 & 16 & 0 & 4 & 18 & 4 & 21\\
\specialrule{\lightrulewidth}{1pt}{1pt}
\multicolumn{9}{@{}l}{\textit{WorldPlay2$^\dagger$}~\cite{worldplay2}}\\
\hspace{4pt}Raw model & 16 & 16 & 16 & 12 & 6 & 16 & 6 & 26\\
\hspace{4pt}Text prompt & 14 & 10 & 11 & 6 & 6 & 12 & 2 & 22\\
\hspace{4pt}External depth & 14 & 10 & 17 & 4 & 5 & 10 & 7 & 21\\
\rowcolor{ours}\hspace{4pt}Ours (Coll) & 10 & 6 & 4 & 2 & 2 & 6 & 2 & 18\\
\rowcolor{ours}\hspace{4pt}Ours (Pred) & 14 & 8 & 7 & 8 & 1 & 4 & 5 & 22\\
\specialrule{\lightrulewidth}{1pt}{1pt}
\multicolumn{9}{@{}l}{\textit{Hy-Nomas$^\dagger$}~\cite{hynomas}}\\
\hspace{4pt}Raw model & 62 & 54 & 43 & 40 & 52 & 62 & 39 & 60\\
\hspace{4pt}Text prompt & 51 & 58 & 43 & 44 & 54 & 60 & 46 & 60\\
\hspace{4pt}External depth & 35 & 40 & 23 & 32 & 42 & 56 & 25 & 50\\
\rowcolor{ours}\hspace{4pt}Ours (Coll) & 39 & 32 & 22 & 20 & 37 & 52 & 30 & 32\\
\rowcolor{ours}\hspace{4pt}Ours (Pred) & 47 & 52 & 33 & 36 & 39 & 54 & 27 & 46\\
\bottomrule
\end{tabular}
\end{table}

\subsection{Task~3 by judge}
\label{s:l3-judges}
This subsection gives Task~3, image-to-video generation from open-domain images, by judge. Four VLM judges score every Task~3 video, Claude Opus~5.5, GPT-6 Astra, DeepSeek V4.1 Flash and Qwen3.8-27B, each with the same frozen prompt. The main paper's Task~3 columns are the mean of the two judges closest to people, Claude and GPT-6, over the static and over the dynamic images; \cref{s:tab-l3-compact} gives each judge on its own.

\subsection{Qualitative results}
\label{s:qual}
\label{s:qual-l2}

\begin{figure}[!t]

\centering
\captionsetup{skip=2.5pt}
\includegraphics[width=0.80\textwidth]{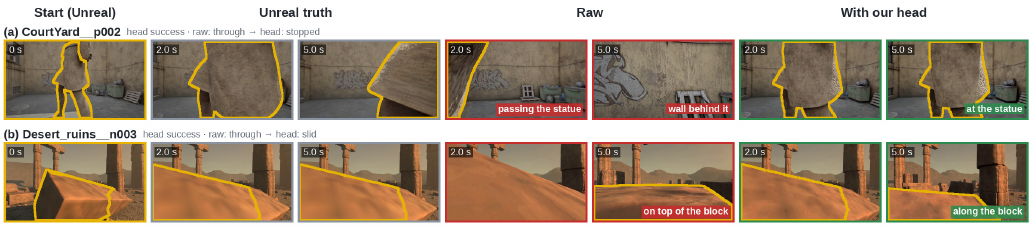}
\caption{\textbf{LingBot-World~2.0} on Task~2 routes (yellow: target obstacle tracked by SAM~2 from hand-off). (a) The raw camera passes through the statue to the wall behind it; with NoPeWorld it stops cleanly in front of the statue [through\,$\to$\,stopped]. (b) The raw camera clips onto the fallen block; with NoPeWorld it slides smoothly along the block towards the pillar [through\,$\to$\,slid].}
\label{s:fig-qual-lingbot}

\vspace{4pt}
\includegraphics[width=0.80\textwidth]{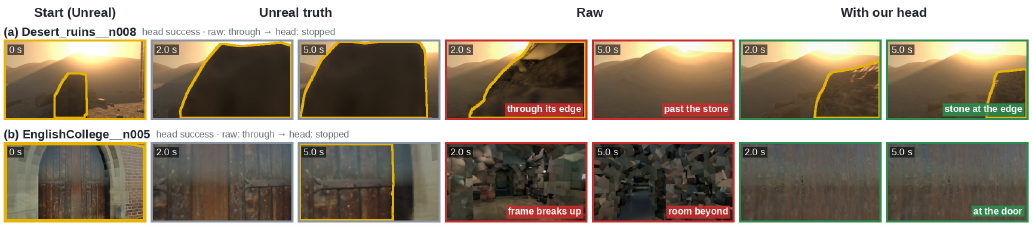}
\caption{\textbf{HY-WorldPlay}, layout as in \cref{s:fig-qual-lingbot}. (a) The raw camera cuts through the rock's edge into open desert; with NoPeWorld it halts beside the rock [through\,$\to$\,stopped]. (b) The raw view fractures after crossing the closed door; with NoPeWorld the camera stops cleanly at the wooden door [through\,$\to$\,stopped].}
\label{s:fig-qual-hy}

\vspace{4pt}
\includegraphics[width=0.80\textwidth]{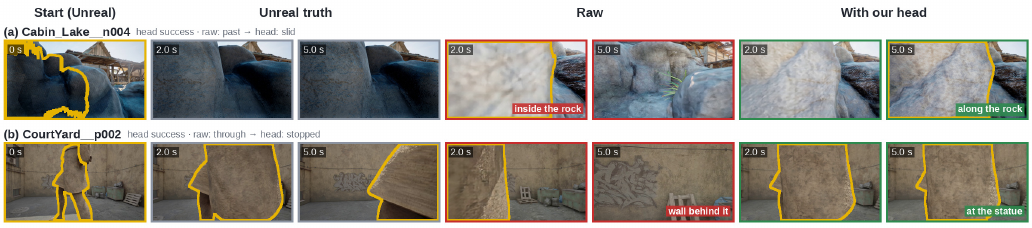}
\caption{\textbf{Echo-WM}, layout as in \cref{s:fig-qual-lingbot}. (a) The raw camera enters the boulder at 2\,s and emerges beyond it at 5\,s; with NoPeWorld it slides along the rock face as in Unreal [past\,$\to$\,slid]. (b) The raw camera clips past the statue's edge to the back wall; with NoPeWorld it stops at the statue [through\,$\to$\,stopped].}
\label{s:fig-qual-echo}

\vspace{4pt}
\includegraphics[width=0.80\textwidth]{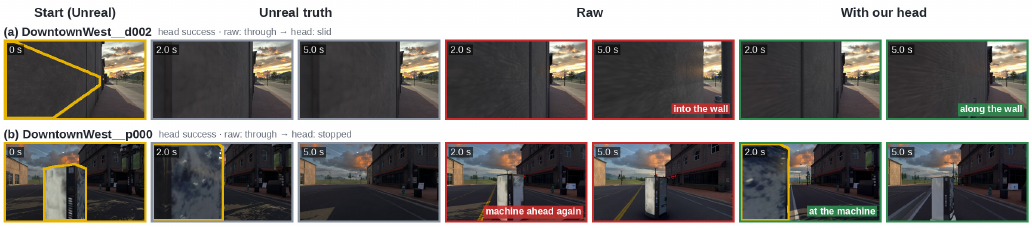}
\caption{\textbf{Evoke-Turbo}, layout as in \cref{s:fig-qual-lingbot}. (a) The raw camera turns into the solid wall; with NoPeWorld it slides along the wall with the street on the right, matching Unreal [through\,$\to$\,slid]. (b) The raw camera clips through the vending machine and resets the view; with NoPeWorld it steers safely past the machine's edge [through\,$\to$\,stopped].}
\label{s:fig-qual-evoke}

\vspace{4pt}
\includegraphics[width=0.80\textwidth]{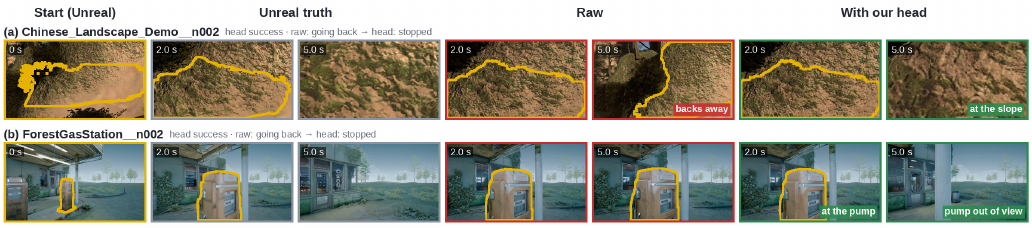}
\caption{\textbf{WorldPlay2}, layout as in \cref{s:fig-qual-lingbot}. (a) The raw camera approaches the slope and retreats backwards after 3.5\,s; with NoPeWorld it holds position at the slope [going back\,$\to$\,stopped]. (b) The raw camera approaches the fuel pump and reverses along its path; with NoPeWorld it reaches the pump and holds without retreating [going back\,$\to$\,stopped].}
\label{s:fig-qual-wp2}

\end{figure}

\Cref{s:fig-qual-lingbot,s:fig-qual-hy,s:fig-qual-echo,s:fig-qual-evoke,s:fig-qual-wp2} show multi-frame Task~2 trajectories across five generators in collision mode from the 100 evaluation routes. Each row begins with the Unreal hand-off frame with the obstacle outlined in yellow (tracked by SAM~2 where visible), followed by Unreal's collide-and-slide reference, the raw model, and +~NoPeWorld at 2.0\,s and 5.0\,s after hand-off (16\,fps). Grey borders denote the reference, red borders mark violations or retreats, and green borders mark safe stops or wall slides (\cref{s:rules}). All frames are unmodified model outputs aside from the obstacle contour; full video rollouts are on the project~page.

\iftasktwomain\else\tasktwofigure\fi

\clearpage
\twocolumn

\begin{table}[!t]
\centering\footnotesize
\caption{\textbf{No future leak} (8 Task~1 routes per generator, replacing all real frames from chunk $k^*$ onward while keeping history, command and seed fixed). $|\Delta P|$: max change in $\Phat_0$ up to $k^*$; Floor: two-run baseline on original frames; Bit-equal: identical-score decisions up to $k^*$; Moved: changed decisions after $k^*$.}
\label{s:tab-noleak}
\setlength{\tabcolsep}{3.5pt}
\resizebox{\columnwidth}{!}{%
\begin{tabular}{@{}llccccc@{}}
\toprule
 & Mode & Decisions & $|\Delta P|$ & Floor & Bit-equal & Moved\\
\midrule
LingBot-World 2.0 & Collision & 67 & 0 & 0 & 67/67 & 47/47\\
 & Prediction & 59 & 0 & 0 & 59/59 & 47/47\\
HY-WorldPlay & Collision & 72 & 0 & 0 & 72/72 & 54/54\\
 & Prediction & 64 & 0 & 0 & 64/64 & 54/54\\
Echo-WM & Collision & 52 & 0 & 0 & 52/52 & 31/31\\
 & Prediction & 44 & 0 & 0 & 44/44 & 31/31\\
Evoke-Turbo & Collision & 30 & 0 & 0 & 30/30 & 17/17\\
WorldPlay2 & Collision & 54 & 0 & 0 & 54/54 & 32/32\\
 & Prediction & 46 & 0 & 0 & 46/46 & 32/32\\
Hy-Nomas & Collision & 64 & 0 & 0 & 64/64 & 54/54\\
 & Prediction & 64 & 0 & 0 & 64/64 & 54/54\\
\bottomrule
\end{tabular}}
\end{table}

\begin{table}[!t]
\centering\footnotesize
\caption{\textbf{How far ahead contact can be read} on LingBot-World~2.0 (AUC of ``input unsafe'' across the four trained heads on 8 validation scenes; positive/negative states in brackets). Lookup: per-chunk training unsafe rate; stratum lookup: chunk-index lookup fitted per lead.}
\label{s:tab-probes}
\setlength{\tabcolsep}{3pt}
\begin{tabular}{@{}lccc@{}}
\toprule
Input unsafe & AUC & Lookup & Stratum lookup\\
\midrule
This chunk (162/1796) & .955--.967 & .69 & .61\\
1 chunk ahead, 0.75\,s (58/1796) & .831--.838 & .59 & .64\\
2 chunks ahead, 1.5\,s (52/1796) & .636--.673 & .48 & .69\\
\bottomrule
\end{tabular}
\end{table}

\section{Analysis}
\label{s:analysis}

\subsection{The head reads no future frame}
\label{s:noleak}
A head decides chunk~$k$ before the chunk is drawn, from the history up to chunk~$k-1$, the command and the noise. \Cref{s:tab-noleak} tests this on Task~1 for every generator and mode: on each route, all real frames from a decision chunk $k^*$ on are replaced by another route's, while history, command and seed stay fixed, and the frozen heads are run again. In every run of \cref{s:tab-noleak}, every decision up to $k^*$ keeps bit-identical logits, and every decision after $k^*$ changes, so the replacement reached the model.

\subsection{How far ahead the contact is read}
\label{s:probe-results}
\label{s:probe-method}
On LingBot-World~2.0, we evaluate the trained heads (\cref{sec:head}) on validation videos rendered in Unreal with ground-truth geometry labels. On full validation videos, the heads predict upcoming contact on the current chunk at .955--.967 AUC and one chunk (0.75\,s) ahead, before its frames are denoised, at .831--.838, far above the training-rate lookup (.59) and stratum lookup (.64; \cref{s:tab-probes}). Two chunks ahead (1.5\,s), AUC drops to .636--.673, matching the stratum lookup (.69), with both modes performing comparably across leads.

\begin{figure}[!t]
\centering
\captionsetup{skip=3pt}
\includegraphics[width=0.67\columnwidth]{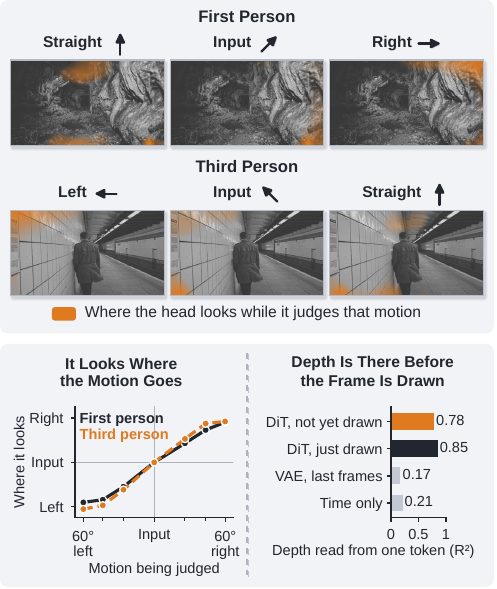}
\caption{\textbf{Where the head looks} (LingBot-World~2.0, collision mode). Top: attention maps (orange) for three candidate motions at first intervention. Bottom left: attention shift across 50 first-person and 40 third-person Unreal routes. Bottom right: single-token depth probe ($R^2$, block 16) vs.\ VAE latents and time.}
\label{fig:interp}
\end{figure}

\begin{table}[!t]
\centering\footnotesize
\caption{\textbf{What the head needs} (LingBot-World~2.0 validation, block 39, 2,378 states). ``Rank'': same-state AUC of safe over unsafe candidates; top two rows read VAE latents or decoded pixels of the preceding chunk (mean of three seeds).}
\label{s:tab-ablation}
\setlength{\tabcolsep}{2.2pt}
\begin{tabular}{@{}lrcccc@{}}
\toprule
Input or head & Params & AUC & \makecell{Full\\video} & Rank & Regret\\
\midrule
Same head on VAE latents & 1.09M & .847 & .816 & .755 & .1101\\
Same head on pixels & 1.47M & .583 & .550 & .647 & .1129\\
Chunk index only & 0 & .79 & .69 & .766 & .1255\\
\midrule
Full head, current chunk & 1.72M & .968 & .958 & .786 & .0554\\
Full head, previous commit & 1.74M & .969 & .949 & .846 & .0561\\
Narrow: 32-d, 1 block & 0.34M & .954 & .928 & .781 & .0584\\
No attention & 1.53M & .954 & .936 & .809 & .0676\\
State, action late merge & 2.05M & .971 & .956 & .712 & .0740\\
\midrule
Always keep the input & 0 & \na & \na & \na & .1128\\
\bottomrule
\end{tabular}
\end{table}
\subsection{What the head needs}
\label{s:reads}
\Cref{s:tab-ablation} compares what each part of the head adds. On LingBot, validation figures come from 8 scenes (\emph{full-video} excludes near-wall clips). Each simplification hurts something different: without attention, misses rise by 0.16; merging state and action only at the end leaves AUC intact but drops same-state ranking by 0.07, so the head can still see that the input is unsafe but cannot judge the alternatives.

\subsection{Where the head looks}
\label{s:where}
We inspect cross-attention inside the LingBot-World~2.0 head (\cref{fig:interp}) on regenerated Task~3 cases (at the first intervention chunk) and on teacher-forced Unreal routes with per-token depth and obstacle masks. Because each candidate motion conditions the token grid before query readout (\cref{eq:head}), attention tracks the queried trajectory: wall-bound motions attend to the blocking surface, forward motions attend to the path ahead, and rightward or leftward turns attend to the corresponding side of the frame.

\end{document}